\documentclass[10pt,twocolumn,letterpaper]{article}
\usepackage[accsupp]{axessibility}

\usepackage{cvpr}              % To produce the CAMERA-READY version

\definecolor{cvprblue}{rgb}{0.21,0.49,0.74}
\usepackage[pagebackref,breaklinks,colorlinks,allcolors=cvprblue]{hyperref}

\usepackage{url}
\usepackage{multicol}
\usepackage{multirow}
\usepackage{array}
\usepackage{wrapfig}
\usepackage{graphicx}
\usepackage{colortbl}
\usepackage[table,dvipsnames]{xcolor}
\usepackage{cuted}
\usepackage{amsmath, mathtools}
\usepackage{enumitem}
\usepackage{subcaption}
\usepackage{pifont}
\usepackage{algorithm}
\usepackage{algpseudocode}
\usepackage{booktabs} 
\usepackage{fontawesome}
\usepackage{bm}
\usepackage{tabularx}
\usepackage{dsfont}
\usepackage{svg}
\definecolor{OursBlue}{RGB}{224,242,255}

\newcommand{\gu}[1]{{\color{green}}}

\newcommand{\cmark}[1][ForestGreen]{\textcolor{#1}{\ding{51}}} % ✓
\newcommand{\xmark}[1][BrickRed]{\textcolor{#1}{\ding{55}}}    % ✗
\newcommand{\Yes}{\cmark}                
\newcommand{\No}{\xmark}                  

\definecolor{OursBlue}{RGB}{224,242,255}  
\definecolor{BestBG}{RGB}{222,255,222}    
\definecolor{SecondBG}{RGB}{255,246,207}

\newcommand{\na}{\textcolor{black!50}{N/A}}  
\newcounter{corrauth}

\newcolumntype{Y}{>{\hspace{6pt}}X<{\hspace{6pt}}}
\newcolumntype{C}{>{\hspace{6pt}}c<{\hspace{6pt}}}

\definecolor{lightblue}{rgb}{0.8,0.9,0.95}

\def\paperID{30069} % *** Enter the Paper ID here
\def\confName{CVPR}
\def\confYear{2026}

\newcommand{\acronym}{InvAD\xspace}
\title{\textit{\acronym}: Inversion-based Reconstruction-Free Anomaly Detection with \\ Diffusion Models}
\author{
Shunsuke Sakai\textsuperscript{1} \quad
Xiangteng He\textsuperscript{2,3} \quad
Chunzhi Gu\textsuperscript{1}\thanks{Corresponding Author.}%
\setcounter{corrauth}{\value{footnote}} \quad
Leonid Sigal\textsuperscript{2,3} \quad
Tatsuhito Hasegawa\textsuperscript{1}\footnotemark[\value{corrauth}] \\
\textsuperscript{1}University of Fukui \quad
\textsuperscript{2}University of British Columbia \quad
\textsuperscript{3}Vector Institute for AI \\
\tt sshunsuke0102@gmail.com, \tt xiangteng.he@ubc.ca, \\ \tt czgu@ieee.org, \tt lsigal@cs.ubc.ca, \tt t-hase@u-fukui.ac.jp
}

\begin{document}
\maketitle

\begin{abstract}
Despite the remarkable success, recent reconstruction-based anomaly detection (AD) methods via diffusion modeling still involve fine-grained noise-strength tuning and computationally expensive multi-step denoising, leading to a fundamental tension between fidelity and efficiency. In this paper, we propose \textbf{InvAD}, a novel {\bf Inv}ersion-based {\bf A}nomaly {\bf D}etection approach – ``\textit{detection via noising in latent space}'' – which circumvents explicit reconstruction. Importantly, we contend that the limitations in prior reconstruction-based methods originate from the prevailing ``\textit{detection via denoising in RGB space}'' paradigm. To address this, we model AD under a reconstruction-free formulation, which directly infers the final latent variable corresponding to the input image via DDIM inversion, and then measures the deviation based on the known prior distribution for anomaly scoring. Specifically, in approximating the original probability flow ODE using the Euler method, we only enforce very few inversion steps to noise the clean image to pursue inference efficiency. As the added noise is adaptively derived with the learned diffusion model, the original features for the clean testing image can still be leveraged to yield high detection accuracy. 
We perform extensive experiments and detailed analysis across four widely used AD benchmarks under the unsupervised unified setting to demonstrate the effectiveness of our model, achieving state-of-the-art AD performance, and about $2\times$ inference time speedup without diffusion distillation. Project page:
\url{https://invad-project.com}
\end{abstract}

\vspace{-13pt}
\section{Introduction}
\label{sec:intro}

%Anomaly Detection%
Image anomaly detection (AD) focuses on identifying images that deviate from normal patterns, with critical applications in industrial defect detection and automated medical diagnosis \cite{Yang2020, bercea2023generalizing}. In real-world scenarios, anomalous samples are often scarce and costly to acquire. Consequently, researchers began focusing on unsupervised AD, where the detection model was trained solely on normal images to identify potential anomalies \cite{PatchCore,EfficientAD}.

\begin{figure}[t]
  \centering
  \includegraphics[width=1.02\linewidth]{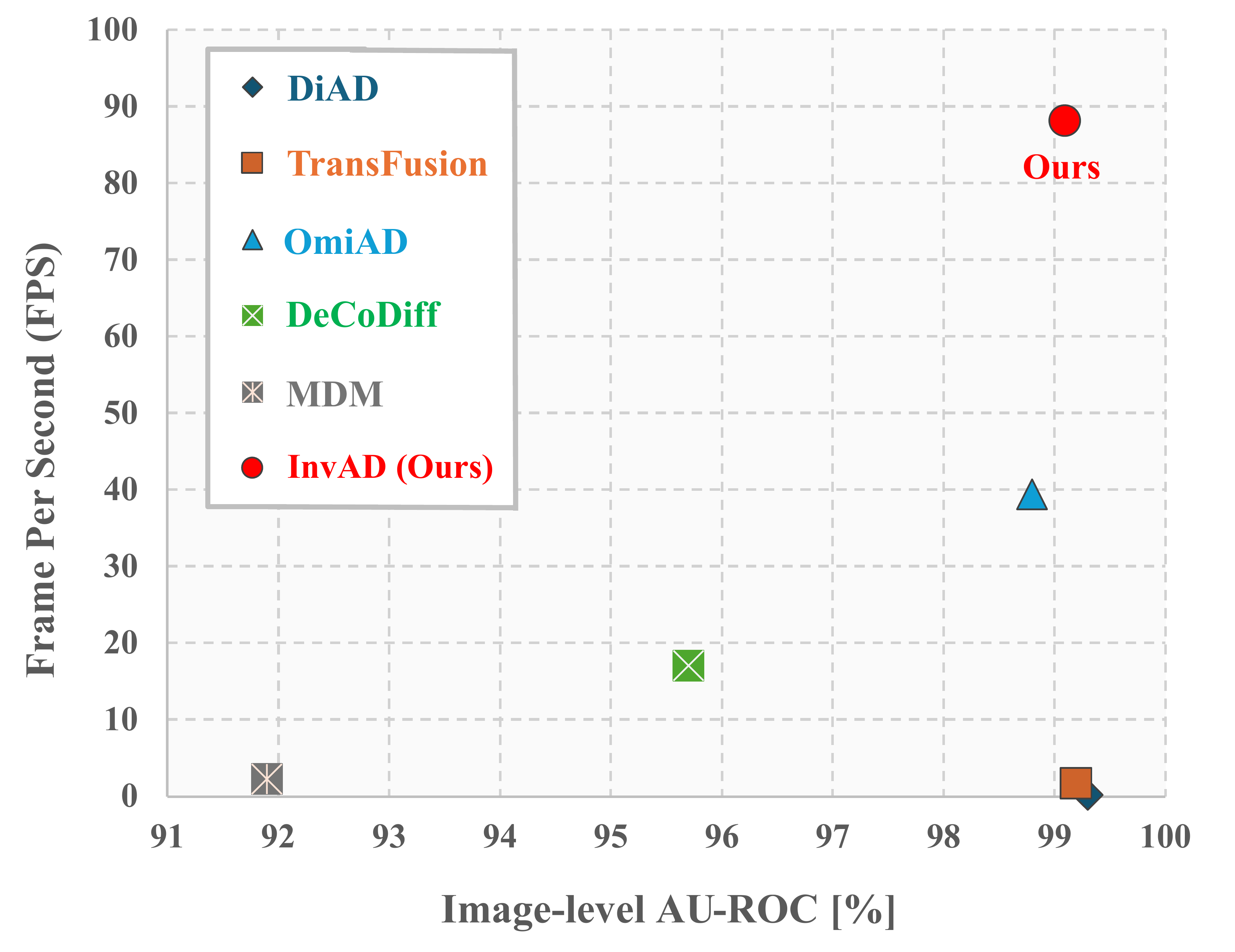}
  \caption{\textbf{Accuracy \textit{v.s.} Speed} relationship of diffusion-based AD methods \cite{DiAD,TransFusion,omiad,decodiff,mad} on MVTecAD. \textit{Our proposed \textcolor{red}{\textbf{InvAD}} achieves state-of-the-art AD performance with a substantial speedup.}}
  \label{fig:scatter_plot_perf}
\end{figure}

\begin{table*}[t]
    \centering
    \caption{\textbf{Comparison of the properties} of diffusion-based AD methods. Normal-only means whether the method involves pseudo-anomalies in training, NFE stands for the number of function evaluations, and TS refers to the timestep.}
    \renewcommand{\arraystretch}{1.2}
    \setlength{\tabcolsep}{3pt}
    \addtolength{\tabcolsep}{3pt}
    \begin{tabularx}{\textwidth}{@{}X c c c c c c@{}}
        \toprule
        \textbf{Method} & \textbf{Normal-only} & \textbf{NFE} & \textbf{FPS} & \textbf{TS Tuning-free} & \textbf{Multi-class} & \textbf{Scoring scheme} \\
        \midrule
        % \textbf{ScoreDD} \textit{\fontsize{7}{10}\selectfont[ArXiv'22]} & \No & $15$ & N/A & \No & \No & $||\nabla_\mathbf{x}p_\theta(\mathbf{x}_t)||_2$ \\
        \textbf{DiffAD} \textit{\fontsize{7}{10}\selectfont[ICCV'23]} & \Yes & \na & \na & \No & \No & Mask prediction \\
        \textbf{DiAD} \textit{\fontsize{7}{10}\selectfont[AAAI'24]} & \No & $10$ & 1.5 & \Yes & \Yes & $\text{MSE}(g_\phi(\mathbf{x}_0), g_\phi(\hat{\mathbf{x}}_0))$\\
        \textbf{GLAD} \textit{\fontsize{7}{10}\selectfont[ECCV'24]} & \Yes & $750$ & 0.2 & \Yes & \No & $\text{MSE}(g_\phi(\mathbf{x}_0), g_\phi(\hat{\mathbf{x}}_0))$ \\
        \textbf{TransFusion} \textit{\fontsize{7}{10}\selectfont[ECCV'24]} & \Yes & $20$ & 1.6 & \No & \No & Mask prediction \\
        % \textbf{DAD} \textit{\fontsize{7}{10}\selectfont[TPAMI'25]} & \Yes & $\{2, 400\}$ & 9.1 & \No & \No & Mask prediction \\
        % \textbf{DDAD} \textit{\fontsize{7}{10}\selectfont[GCPR'24]} & \No & $\{5, 10, 25\}$ & 0.8 & \No & \No & $\text{MSE}(g_\phi(\mathbf{x}_0), g_\phi(\hat{\mathbf{x}}_0))$ \\
        \textbf{MDM} \textit{\fontsize{7}{10}\selectfont[ICML'25]} & \Yes & $40$ & 1.9 & \No & \No & $\text{MSE}(g_\phi(\mathbf{x}_0), g_\phi(\hat{\mathbf{x}}_0))$ \\
        \textbf{OmiAD} \textit{\fontsize{7}{10}\selectfont[ICML'25]} & \Yes & $1$ & 39.4 & \No & \Yes & $\text{MSE}(g_\phi(\mathbf{x}_0), g_\phi(\hat{\mathbf{x}}_0))$ \\
        \textbf{DeCo-Diff} \textit{\fontsize{7}{10}\selectfont[CVPR'25]} & \Yes & $10$ & 17.0 & \No & \Yes & $\text{MSE}(g_\phi(\mathbf{x}_0), g_\phi(\hat{\mathbf{x}}_0))$ \\
        \midrule
        \rowcolor{OursBlue}
        \textbf{InvAD (Ours)} & \textbf{\Yes} & \textbf{3} & \textbf{88.1} & \textbf{\Yes} & \textbf{\Yes} & $\log p_\theta(\mathbf{x}_T)$ \\
        \bottomrule
    \end{tabularx}
    \addtolength{\tabcolsep}{-3pt}
    \label{tab:diffusion_comparison}
\end{table*}

\begin{figure*}[!t] % t は top (上部配置)
    \centering
    \begin{subfigure}{0.49\textwidth}
        \centering
        \includegraphics[width=\linewidth]{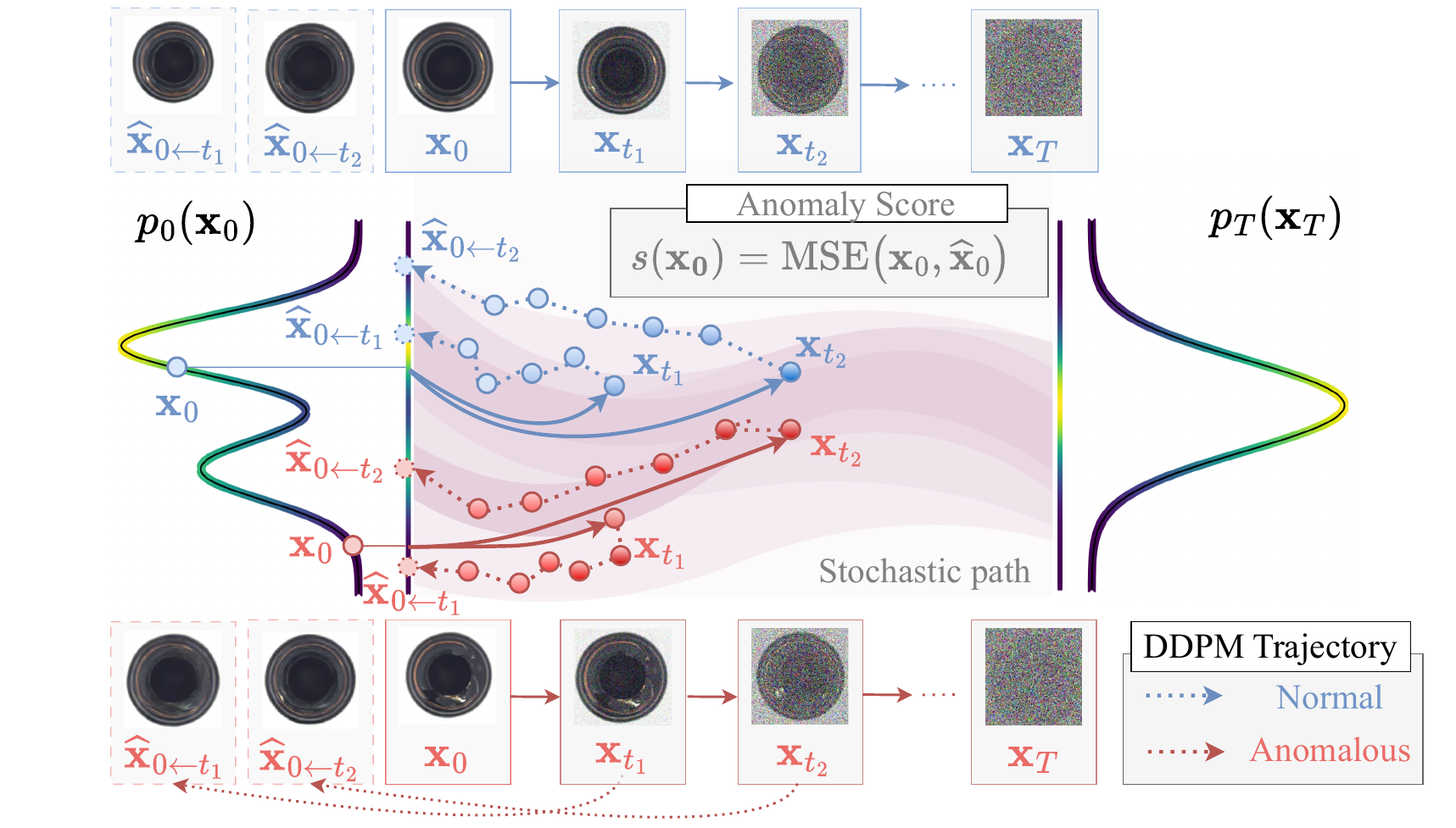}
        \caption{Conventional reconstruction-based AD paradigm}
        \label{fig:method_recon}
    \end{subfigure}
    \hfill
    \begin{subfigure}{0.49\textwidth}
        \centering
        \includegraphics[width=\linewidth]{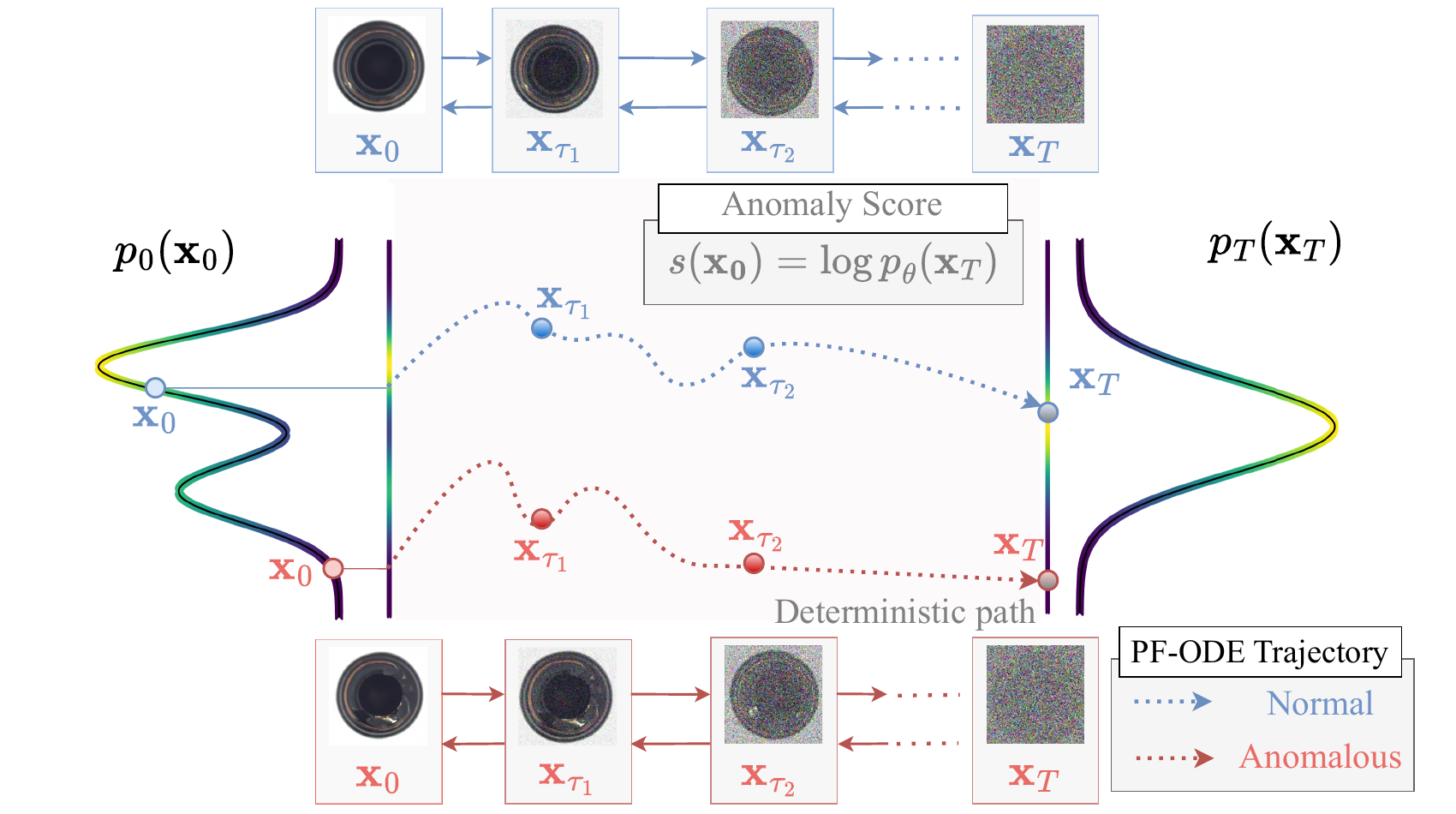}
        \caption{Our inversion-based AD paradigm}
        \label{fig:method_ours}
    \end{subfigure}
    \caption{\textbf{Conceptual comparison} of conventional and our proposed AD paradigm. Conventional reconstruction-based paradigm (a) %\cite{DiffAD,GLAD,RemoveAsNoise,DiffusionAD,DDAD,DiAD} 
    first perturbs an input sample \(\mathbf{x}_0\) to a latent state \(\mathbf{x}_t\) at step \(t\)
    %via equation \ref{eq:q_prop}
    , and then denoises \(\mathbf{x}_t\) back to \(\mathbf{x}_0\) %using equation \ref{eq:generative_process1}
    . The anomaly score is computed as the mean squared error (MSE) between the original input and its reconstructed sample. In contrast, our inversion-based paradigm (b) directly infers the latent state at the final step, \(\mathbf{x}_T\), by tracing the PF-ODE trajectories. The anomaly score is then determined based on the typicality of \(\mathbf{x}_T\) within the tractable latent distribution.}
    \label{fig:method_overview}
\end{figure*}

%Diffusion-based AD%
Due to the robust capabilities for approximating distributions, diffusion models \cite{DDPM,DiffusionModel,dhariwal2021diffusion} have emerged as powerful tools for AD by precisely modeling normal data distribution \cite{DiffAD,GLAD,RemoveAsNoise,DiffusionAD,DDAD,DiAD,AnoDDPM,TransFusion,ScoreDD,omiad,decodiff,dte,vpot}. 
The prevailing paradigm in these methods is to treat anomalies as noise: by adding noise of a certain intensity to an anomalous image and then denoising it with a diffusion model trained solely on normal data, the anomalous regions can be effectively reconstructed as normal regions. The anomaly score is therefore measured with the mean squared error (MSE) between the original input and its reconstructed counterpart, as shown in \Cref{fig:method_overview} (a). 

We argue that the above ``\textit{detection via denoising in RGB space}'' paradigm is less effective given the following fundamental limitations: \textit{(1) Noise-strength sensitivity}. The detection performance depends heavily on the selection of appropriate noise strengths. Stronger noise unnecessarily corrupts normal regions, increasing false positives, while weaker noise leads to overly well-recovered anomaly regions
\cite{RemoveAsNoise,GLAD}; \textit{(2) Computational expense in multi-step denoising}. Satisfactory reconstruction via diffusion models often requires iterative denoising, which is time-consuming for real-world applications. As listed in \Cref{tab:diffusion_comparison}, most reconstruction-based methods involve high inference latency. %\mumu{The results should be consistent with those in Table 5.}

Therefore, we introduce the “\textit{detection via noising in latent space}" paradigm, which bypasses the reconstruction process in prior AD approaches. Our core insight is that, since the learned diffusion model is solely aware of the normal data, we can utilize such learned normal data distribution through latent inversion, rather than reconstruction, as shown in \Cref{fig:method_overview} (b).  Specifically, for a given image, we directly infer the corresponding latent variable at the final diffusion step by tracing the probability flow ordinary differential equation (PF-ODE \cite{SBGM}) trajectory. Since PF-ODE is by nature deterministic, this constructs a one-to-one 
correspondence between the noise prior distribution and the distribution of normal images, ensuring that normal images can be ideally mapped to focus on a high-density region. Conversely, anomalous images will be mapped to low-density regions. Hence, we can detect anomalies by evaluating the degree of deviation of the inferred latent variable under the prior distribution. 

%提案手法の課題%
While the above method detours noise strength tuning, it does not straightforwardly lead to a reduction of detection speed. In particular, the high inference latency primarily stems from the high-order ODE solver introduced for accurate reconstruction from inverted latents. To address this, we are inspired by \cite{ScoreDD,diffpath} to measure the norm of the score function as a proxy for typicality, which is robust to the step size selection of the Euler method. This allows us to enforce very few inversion steps to diffuse the clean image to pursue inference efficiency. Moreover, it contributes to a significantly more efficient inference paradigm compared to prior approaches. Importantly, because the learned diffusion model enables noise to be adaptively added during inversion, the key features of the original clean testing image can still be leveraged to harvest high
detection accuracy. As plotted in \Cref{fig:scatter_plot_perf}, our InvAD achieves state-of-the-art detection performance with the inference efficiency surpassing prior diffusion-based AD models by a large margin. Also, as our method is designed for the inference stage, it can be applied in a plug-and-play fashion to be incorporated into existing AD models. Experimental evaluations demonstrate that our method substantially improves inference latency, yet still yields high AD performance, across four widely-used AD benchmarks.

%Contribution%
Our core contributions include three folds:
  
\begin{itemize}
\item 
Identifying the inherent limitations of existing reconstruction-based methods in achieving an optimal efficiency-accuracy balance, which we attribute fundamentally to the ``detection via denoising in RGB space'' paradigm.
\item 
Proposing a novel paradigm, detection via noising in latent space, with an inversion-based AD framework \textbf{InvAD} that eliminates reconstruction requirements by directly estimating latent variables from inputs and quantifying anomaly scores through deviation measurements under the prior distribution.
\item 
Demonstrating that our inversion-based approach, which infers latent variables, is simpler, faster, and achieves state-of-the-art detection performance compared to conventional reconstruction-based methods.
\end{itemize}

\section{Background}
\label{sec:background}

\subsection{Diffusion Models}
\label{sec:diffusion}

Diffusion modeling \cite{DDPM,SBGM} is a recently emerged generative modeling framework, which aims to reverse a Markov perturbation process (\textit{forward} process). In general, each transition of the \textit{forward} process is defined as:
\begin{align}
q\bigl(\mathbf{x}_t \mid \mathbf{x}_{t-1}\bigr)
&:= \mathcal{N}\!\Bigl(
    \sqrt{\frac{\alpha_t}{\alpha_{t-1}}}\,\mathbf{x}_{t-1},
    \bigl(1 - \frac{\alpha_t}{\alpha_{t-1}}\bigr)\mathbf{I}
\Bigr),
\label{eq:forward_process}
\end{align}
where the monotonically decreasing sequence \(\alpha_{1:T} \in (0, 1]^T\) schedules the noise intensity. The sampling procedure can then be achieved with a
 \textit{reverse} process, which defines a Markov process with a Gaussian transition kernel:
\begin{align}
p_\theta(\mathbf{x}_{t-1} \mid \mathbf{x}_t) := \mathcal{N}\bigl(\mathbf{x}_{t-1}; \boldsymbol{\mu}_\theta^{(t)}(\mathbf{x}_t),  \bigl(1 - \frac{\alpha_t}{\alpha_{t-1}}\bigr)\mathbf{I}\bigr),
\label{eq:generative_process2}
\end{align}
where $\theta$ denotes the learnable parameter of the DDPM. The training is realized by minimizing the $\epsilon$-prediction loss:
\begin{equation}
    \mathcal{L}_\gamma\left(\epsilon_\theta\right) \coloneqq \sum_{t=1}^T \gamma_t \mathbb{E}_{\mathbf{x}_0 \sim q\left(\mathbf{x}_0\right), \epsilon_t \sim \mathcal{N}(\mathbf{0}, \mathbf{I})}\left[\left\|\epsilon_\theta^{(t)}(\mathbf{x}_t)-\epsilon_t\right\|_2^2\right],
    \label{eq:diffusion_objective2}
\end{equation}
where $\mathbf{x}_t$ is efficiently sampled from $q_t(\mathbf{x}_t\mid \mathbf{x}_0)=\mathcal{N}(\mathbf{x}_t; \sqrt{\alpha_t}\mathbf{x}_0, (1-\alpha_t)\mathbf{I})$, and $\mathbf{\gamma} \coloneqq [\gamma_1, \ldots, \gamma_T]^T$ denotes the weighting vector. During sampling, one can first sample a pure noise from $p_\theta(\mathbf{x}_T) \approx \mathcal{N}(\mathbf{0}, \mathbf{I})$, and then progressively trace \Cref{eq:generative_process2} for denoising. 

\subsection{Denoising Diffusion Implicit Models}
The Gaussian approximation in \Cref{eq:generative_process2} generally works satisfactorily when the total number of diffusion steps, \(T\), is sufficiently large \cite{DDPM}. However, increasing \(T\) also magnifies the number of function evaluations (NFE), which slows down the sampling procedure. To address this, Song \textit{et al.} introduced denoising diffusion implicit models (DDIM), allowing for faster sampling by leveraging a subset of the original diffusion process \cite{DDIM}. Specifically, they reformulate the \textit{reverse} process as follows:
\begin{equation}
\resizebox{\columnwidth}{!}{
  $\displaystyle
    \begin{aligned}
       \mathbf{x}_{\tau_{i\!-\!1}} &= \sqrt{\alpha_{\tau_{i\!-\!1}}}
       f_\theta(\mathbf{x}_{\tau_{i}})
       + \sqrt{1-\alpha_{\tau_{i\!-\!1}}}\,\epsilon_\theta^{(\tau_{i})}\bigl(\mathbf{x}_{\tau_{i}}\bigr),
    \end{aligned}
  $
}
\label{eq:ddim_sampling3}
\end{equation}
where $f_\theta(\mathbf{x}_{\tau_{i}})
       = \mathbf{x}_{\tau_{i}} - \sqrt{1-\alpha_{\tau_{i}}}\,\epsilon_\theta^{(\tau_{i})}\bigl(\mathbf{x}_{\tau_{i}}\bigr)/\sqrt{\alpha_{\tau_{i}}}$ and  $\tau_S = [\tau_1, \tau_2, \ldots, \tau_S = T] \subset \{1, 2, \ldots, T\}$. This deterministic sampling equation can be interpreted as the Euler integration of the corresponding PF-ODE:
\begin{equation}
    \mathrm{d}\mathbf{y}_t = \epsilon_\theta^{(t)} \,\mathrm{d}p_t,
    \label{eq:ddim_ode}
\end{equation}
where \(\mathbf{y}_t \coloneqq \mathbf{x}_t/\sqrt{\alpha_t}\) and \(p_t \coloneqq \sqrt{1/{\alpha_t} - 1}\). The PF-ODE gradually transforms a noise distribution into a data distribution, and vice versa. This approach, known as  \textit{inversion}, maps
a given data sample back to its noise representation.

%In doing so, it provides a method—referred to as \textit{inversion}—to map a given data sample back to its corresponding noise representation.

\subsection{Diffusion-Based Anomaly Detection}
The above diffusion-based generative models have been recently employed to handle AD. Existing diffusion-based AD methods generally follow the reconstruction-based paradigm. In particular, to address the challenge of noise level selection inherent in this framework, these approaches can be broadly classified into the following three categories:

\begin{itemize}
    \item \textit{Conditional information injection}-based methods incorporate auxiliary information, such as class labels \cite{DiAD}, semantic features \cite{DDAD,DiAD,vpot}, or unmasked regions \cite{omiad,maediff,mad}. As pointed out in \cite{DDAD}, conditioning the denosing process with such additional information contributes to reducing reconstruction error in the normal region and increasing it for anomalous regions. However, since this line of methods conditions the diffusion model at training time, they can lead to increased training complexity.

    \item \textit{Discriminator}-based methods introduce an additional discriminator, typically trained to distinguish between normal and anomalous pixels \cite{DiffAD,DiffusionAD}. While this line of methods enables efficiency, they rely on pseudo-anomalies, biasing AD towards specific anomaly types. 

    \item \textit{Adaptive timesteps}-based methods attempt to adjust or identify perturbation timesteps by comparing the reconstruction errors of denoised results across different time steps \cite{GLAD}. 
    As the above two types of methods do not incorporate an explicit mechanism for adjusting the perturbation time step, it is typically treated as a hyperparameter or addressed via ensembling over multiple time steps \cite{RemoveAsNoise}. See the 5th column in \Cref{tab:diffusion_comparison} for more details.
\end{itemize}

\label{sec:diffusion_based_ad}
In contrast to the state-of-the-art non-diffusion method \cite{EfficientAD} that achieves over 200 FPS, current diffusion-based methods mostly reach 1 FPS on average. While a recent diffusion-based approach \cite{omiad} tackles this by introducing score distillation to constrain the diffusion iteration to only one step, the training for distillation can be complicated due to the instability of the adversarial learning. \Cref{tab:diffusion_comparison} summarizes the properties of current diffusion-based AD methods. Overall, a simple, efficient, yet effective paradigm for diffusion-based AD remains unexplored to date. 
\section{Methodology}
\label{sec:method}

%\subsection{Overview}
%\label{sec:method_overview}
To address the limitations of current diffusion-based methods, we introduce a novel diffusion-based AD framework with \textit{latent inversion}, where the sampling trajectory is reformulated as \(\mathbf{x}_0 \rightarrow \mathbf{x}_T\). In contrast to conventional reconstruction-based approaches \(\bigl(\mathbf{x}_0 \rightarrow \mathbf{x}_t \rightarrow \hat{\mathbf{x}}_0\bigr)\), our method infers the final latent state \(\mathbf{x}_T\) directly from \(\mathbf{x}_0\) by following the \textbf{learned} PF-ODE trajectory. \Cref{fig:method_overview}(b) illustrates our reconstruction-free AD paradigm, which leverages latent inversion to achieve efficient yet accurate AD.

\subsection{Inversion-Based Anomaly Detection}
\label{sec:inversion_ad}
%How to infer corresponding latent, reverse manner%
Given an image \(\mathbf{x}\) and a score function \(\epsilon_\theta^{(t)}\) trained using \Cref{eq:diffusion_objective2}, our goal is to infer the latent representation \(\mathbf{x}_T\) corresponding to \(\mathbf{x}\) by tracing the PF-ODE trajectory. In this work, we employ the PF-ODE derived from the DDIM framework \cite{DDIM}, as shown in \Cref{eq:ddim_ode}. By applying an Euler approximation to \Cref{eq:ddim_ode} and integrating forward, we obtain the following discrete update equation for latent inversion:
\begin{align}
\mathbf{x}_{t+1} &= \sqrt{\alpha_{t+1}}
    f_\theta(\mathbf{x}_t)
    + \sqrt{1-\alpha_{t+1}}\,\epsilon_\theta^{(t)}\bigl(\mathbf{x}_t\bigr).
\label{eq:ddim_inversion1}
\end{align}
However, directly deriving $\mathbf{x}_T$ with \Cref{eq:ddim_inversion1} involves iterative computation of $\epsilon_\theta^{(t)}$ for $T$ times, leading to slower inference. Here, we propose to only use the partial process of the original diffusion process in \Cref{eq:forward_process}. Given the discrete timestep set
$\{1, 2, \ldots, T\}$, we prepare a monotonically increasing subset $\bm{\tau}_S = [\tau_1, \tau_2, \ldots, \tau_S=T] \subset \{1, 2, \ldots, T\}$, where $S \in \{1, 2, \ldots, T\}$ is the number of the partial process. The inversion modeling can thus be accelerated by simply noising the test image with fewer timesteps stored in $\bm{\tau}_S$, which is given by:
\begin{align}
\mathbf{x}_{\tau_{i+1}} &= \sqrt{\alpha_{\tau_{i+1}}}
    f_\theta(\mathbf{x}_{\tau_{i}})
    + \sqrt{1-\alpha_{\tau_{i+1}}}\,\epsilon_\theta^{(\tau_{i})}\bigl(\mathbf{x}_{\tau_{i}}\bigr).
\label{eq:ddim_inversion2}
\end{align}
Intuitively, \Cref{eq:ddim_inversion2} leads to a considerable speed-up for the inference stage when $T \gg S$. Nevertheless, it also inevitably sacrifices approximation precision as the step interval for the Euler integration is extended. \textit{Recall here that our goal is to resolve AD rather than pursue accurate pixel-wise reconstruction.} As such, we argue that high AD performance does not necessarily require reconstruction precision. It should be noticed that, even with limited inversion steps, the inversion mechanism in projecting an anomaly pixel can nonetheless yield sufficiently low typicality. In this regard, despite the poorly reconstructed samples, an anomaly pixel can still be expectedly mapped to a low-density region within the noise prior to achieve high detection performance. 
This insight suggests that our inversion-based AD method is free from the accuracy-efficiency trade-off in conventional reconstruction-based methods. We next need to find an ideal space that best suits our diffusion-based AD modeling. See \textit{Supp. Mat.} for more detailed derivations of inversion.

\subsection{Diffusion Modeling in Feature Space}
\label{sec:feature_diffusion}
Current image AD methods \cite{EfficientAD,PatchCore,FastFlow,DDAD} heavily rely on pre-trained networks, referred to as \textit{backbones}, that are trained on large-scale datasets such as ImageNet \cite{ImageNet}. In diffusion-based AD, these backbones are used to compare the original image with its reconstructed counterpart in feature space \cite{DDAD,RemoveAsNoise,DiAD}. Because backbone features exhibit invariance to low-level variations (\textit{e.g.}, noise or lighting conditions), they can significantly enhance AD performance.

However, in our inversion-based approach, the inferred latent $\mathbf{x}_T$ follows a standard Gaussian distribution, not the data distribution. To nonetheless utilize backbone models, we design our method to follow the latent-diffusion manner by extracting the feature $\mathbf{z} $ via $\mathbf{z} = g_\phi(\mathbf{x}) \in \mathbb{R}^{C \times h \times w}$, where $g_\phi(\cdot)$ denotes the backbone model, and $C, (h, w)$ mean the feature map dimension and resolutions, respectively. This feature-space diffusion modeling enables our inference stage to refer to high-level semantics, rather than low-level pixel patterns. Also, such a latent-diffusion-based framework leads to further inference efficiency.

\begin{algorithm}[t]
\caption{Inference with InvAD}
\begin{algorithmic}[1]
\Ensure $\epsilon_\theta^{(t)}$, $g_\phi$, $\mathbf{x}_0 \in \mathbb{R}^{3\times H \times W}$, $\tau_S = [\tau_1, \ldots, \tau_S]$, $\alpha_{1:T}$
\Require Anomaly score $s$, Anomaly map $A \in \mathbb{R}^{H \times W}$

\State \textit{/* DDIM Inversion */}
\State Initialize $\mathbf{z} \gets g_\phi(\mathbf{x}_0) \in \mathbb{R}^{C \times h \times w}$
\For{$i = 0$ to $S-1$}
    \State Update $\mathbf{z}$ with Equation ~\ref{eq:ddim_inversion2}
    \If{$i = S - 1$}
        \State \textbf{break}
    \EndIf
\EndFor

\State \textit{/* Anomaly scoring */}
\State Initialize $A \gets \mathbf{0} \in \mathbb{R}^{h \times w}$
\For{$u\in\{0,\ldots,h-1\}, v \in \{0,\ldots,w-1\}$}
\State Update $A[u, v]$ with Equation \ref{eq:scoring_norm}
\EndFor

\State Upsample $A \in \mathbb{R}^{h \times w}$ to $\mathbb{R}^{H \times W} $
\State $s \leftarrow \max(A) - \min(A) + \sum_{u,v} A[u,v]$
\State \Return $(s, A)$

\end{algorithmic}
\label{alg:proposed_method}
\end{algorithm}

\subsection{Anomaly Score Calculation}
\label{sec:anomaly_score}
Once we obtain the latent representation \(\mathbf{z}_T \in \mathbb{R}^{C \times h \times w}\) for features of a given test sample \(\mathbf{x}\), the most straightforward way to compute the anomaly score is via the log-likelihood under the latent Gaussian distribution. However, it often triggers the known 
\textit{reverse-scoring} problem and fails to capture small anomalies (See \textit{Supp. Mat.}). Given that anomalies are inherently sparse in images, we additionally propose a norm-based image-level anomaly scoring approach. Specifically, we compute the Euclidean norm over all channels $C$ for each pixel \((i, j)\) 
in \(\mathbf{z}_T\), which is written as:
\begin{equation}
\mathbf{z}_T^{\text {normed }}[i, j]=\| \mathbf{z}_T[:, i, j]\|_2 \quad.
\label{eq:scoring_norm}
\end{equation}
Different from \cite{DiAD}, for the image-level anomaly score, we compute the difference between the maximum and minimum values of \(\mathbf{z}_T^{\text{normed}}\). This is motivated by the observation that, even if an image is anomalous, anomalies are often locally and sparsely distributed within small regions, and the majority of image regions remain normal. In essence, our scheme for image-level anomaly scoring examines the difference between the highest- and the lowest-norm pixels to mitigate the influence of outliers, thereby alleviating the reverse-scoring issue. The pixel-level anomaly score  \(A\) is simply produced by extending \(\mathbf{z}_T^{\text{normed}} \in \mathbb{R}^{h \times w}\) to match the desired output resolution \((H, W)\) via bilinear interpolation.

Our proposed InvAD paradigm is summarized in \Cref{alg:proposed_method}. It is efficient yet effective, and can be applied in a plug-and-play manner for existing diffusion-based AD models, as it is devised only for the inference stage. 
\section{Experiments}
\label{sec:experiments}

\subsection{Datasets}
\label{sec:datasets}
%rough description of these datasets%
We use three widely-used unsupervised AD benchmarks—MVTecAD \cite{MVTecAD}, VisA \cite{ViSA}, MPDD \cite{MPDD}—all designed for measuring real-world AD performance on RGB industrial images. We also experiment on the BMAD \cite{bmad} benchmark (with six reorganized datasets) to study the effectiveness of InvAD on the medical AD domain. Since each benchmark provides both image- and pixel-level annotations, we can evaluate performance in terms of overall AD (image-level) as well as precise localization (pixel-level). 
See \textit{Supp. Mat.} for a detailed description of these datasets.

\begin{table*}[!t]
\centering
\small
\caption{\textbf{Quantitative results} on different AD datasets under the multi-class setting. The best and the second-best results are highlighted in bold and underlined, respectively. For each row, the mAD averages the evaluation over all image- and pixel-level metrics (\textit{i.e.}, the 3rd to the 9th column). FPS measures the detection efficiency.}
\label{tab:ad_visa_main}
\begin{tabular}{llcccccccccc}
\toprule
\textbf{Dataset} & \textbf{Method} & \multicolumn{4}{c}{\textbf{Image-level}} & \multicolumn{4}{c}{\textbf{Pixel-level}} & \textbf{mAD} & \textbf{FPS} \\
\cmidrule(lr){3-6} \cmidrule(lr){7-10}
& & AU-ROC & AP & F1\_max & & AU-ROC & AP & F1\_max & AU-PRO & & \\
\midrule
\multirow{8}{*}{MVTecAD}
& RD4AD   \textit{\fontsize{7}{10}\selectfont[CVPR'22]}     & 94.6 & 96.5 & 95.2 &      & 96.1 & 48.6 & 53.8 & 91.1 & 82.3 & 4.8 \\
& UniAD   \textit{\fontsize{7}{10}\selectfont[NeurIPS'22]}    & 96.5 & 98.8 & 96.2 &      & 96.8 & 43.4 & 49.5 & 90.7 & 81.7 & 5.3 \\
& SimpleNet  \textit{\fontsize{7}{10}\selectfont[CVPR'23]} & 95.3 & 98.4 & 95.8 &      & 96.9 & 45.9 & 49.7 & 86.5 & 81.2 & \na \\
& DeSTSeg   \textit{\fontsize{7}{10}\selectfont[CVPR'23]}   & 89.2 & 95.5 & 91.6 &      & 93.1 & \textbf{54.3} & 50.9 & 64.8 & 77.1 & \na \\
& DiAD   \textit{\fontsize{7}{10}\selectfont[AAAI'24]}     & 97.2 & 99.0 & 96.5 &      & 96.8 & \underline{52.6} & \underline{55.5} & 90.7 & \underline{84.0} & 0.1 \\
& HVQ-Trans \textit{\fontsize{7}{10}\selectfont[NeurIPS'23]}  & 98.0 & 99.5 & \underline{97.5} &      & 97.3 & 48.2 & 53.3 & 91.4 & 83.6 & 5.6 \\
& MDM \textit{\fontsize{7}{10}\selectfont[ICML'25]}& 87.2 & 94.2 & 91.0 & & 94.8 & 39.5 & 44.8 & 88.0 & 77.1 & 1.9 \\
& OmiAD \textit{\fontsize{7}{10}\selectfont[ICML'25]}& \underline{98.8} & \textbf{99.7} & \textbf{98.5} & & \textbf{97.7} & \underline{52.6} & \textbf{56.7} & \textbf{93.2} & \textbf{85.3} & 39.4 \\
& \textbf{InvAD (Ours)} & \textbf{99.0} & \underline{99.6} & \textbf{98.5} & & \underline{97.5} & 46.5 & 52.3 & \underline{92.7} & 83.7 & \textbf{88.1} \\
\midrule
\multirow{8}{*}{VisA}
& RD4AD   \textit{\fontsize{7}{10}\selectfont[CVPR'22]}    & 92.4 & 92.4 & 89.6 &      & 98.1 & 38.0 & 42.6 & \underline{91.8} & 77.8 & 4.9 \\
& UniAD   \textit{\fontsize{7}{10}\selectfont[NeurIPS'22]}    & 88.8 & 90.8 & 85.8 &      & 98.3 & 33.7 & 39.0 & 85.5 & 74.6 & 4.6 \\
& SimpleNet  \textit{\fontsize{7}{10}\selectfont[CVPR'23]} & 87.2 & 87.0 & 81.8 &      & 96.8 & 34.7 & 37.8 & 81.4 & 72.4 & \na \\
& DeSTSeg   \textit{\fontsize{7}{10}\selectfont[CVPR'23]}  & 88.9 & 89.0 & 85.2 &      & 96.1 & \underline{39.6} & \underline{43.4} & 67.4 & 72.8 & \na \\
& DiAD \textit{\fontsize{7}{10}\selectfont[AAAI'24]}       & 86.8 & 88.3 & 85.1 &      & 96.0 & 26.1 & 33.0 & 75.2 & 70.1 & 0.1 \\
& HVQ-Trans \textit{\fontsize{7}{10}\selectfont[NeurIPS'23]}  & 93.2 & 92.8 & 87.6 &      & 98.7 & 35.0 & 39.6 & 86.3 & 76.2 & 5.0 \\
& OmiAD \textit{\fontsize{7}{10}\selectfont[ICML'25]}& \underline{95.3} & \underline{96.0} & \underline{91.2} & & \underline{98.9} & \textbf{40.4} & \textbf{44.1} & 89.2 & \underline{79.3} & 35.3 \\
& \textbf{InvAD (Ours)} & \textbf{96.9} & \textbf{97.2} & \textbf{93.7} & & \textbf{99.1} & 39.2 & 43.1 & \textbf{92.9} & \textbf{80.3} & \textbf{74.1} \\
\midrule
\multirow{8}{*}{MPDD}
& RD4AD   \textit{\fontsize{7}{10}\selectfont[CVPR'22]}    & 84.1 & 83.2 & 84.1 &      & 98.1 & 35.2 & 38.7 & \underline{93.4} & 73.8 & 4.7 \\
& UniAD   \textit{\fontsize{7}{10}\selectfont[NeurIPS'22]} & 82.2 & 87.1 & 85.1 &      & 95.1 & 18.9 & 25.0 & 81.9 & 67.9 & 5.8 \\
& SimpleNet \textit{\fontsize{7}{10}\selectfont[CVPR'23]}  & 90.6 & 94.1 & 89.7 &      & 97.1 & 33.6 & 35.7 & 90.0 & 75.8 & \na \\
& DeSTSeg   \textit{\fontsize{7}{10}\selectfont[CVPR'23]}  & 93.0 & 95.1 & 90.6 &      & 94.1 & 33.2 & 37.6 & 59.8 & 71.9 & \na \\
& DiAD   \textit{\fontsize{7}{10}\selectfont[AAAI'24]}     & 74.6 & 82.1 & 82.5 &      & 93.0 & 15.9 & 21.2 & 78.4 & 64.0 & 0.1 \\
& HVQ-Trans \textit{\fontsize{7}{10}\selectfont[NeurIPS'23]} & 86.5 & 88.1 & 85.8 &      & 96.7 & 27.6 & 31.4 & 86.9 & 71.9 & 6.2 \\
& OmiAD \textit{\fontsize{7}{10}\selectfont[ICML'25]} & \underline{93.7} & \underline{95.5} & \underline{90.9} & & \textbf{98.6} & \underline{37.6} & \underline{42.3} & \textbf{94.0} & \underline{78.9} & 49.8 \\
& \textbf{InvAD (Ours)} & \textbf{96.5} & \textbf{96.5} & \textbf{94.4} & & \underline{98.3} & \textbf{40.1} & \textbf{43.9} & 91.2 & \textbf{80.1} & \textbf{120} \\
\bottomrule
\end{tabular}
\end{table*}

\begin{table*}[ht]
\small
\centering
\caption{\textbf{Quantitative results} on BMAD under the single-class setting, where we report image-level AU-ROC (det.), pixel-level AU-ROC (loc.), and FPS. The best and the second-best results are highlighted in bold and underlined, respectively. For each row, the mAD averages the evaluation over all image- and pixel-level metrics. \textit{FPS measures the detection efficiency on the RESC dataset.}}
\begin{tabular}{lccccccccccc}
\toprule
\textbf{Methods} & \multicolumn{2}{c}{\textbf{BraTS2021}} & \multicolumn{2}{c}{\textbf{BTCV + LiTS}} & \multicolumn{2}{c}{\textbf{RESC}} & \textbf{OCT2017} & \textbf{RSNA} & \textbf{Camelyon16} & \textbf{mAD} & \textbf{FPS} \\ 
\cmidrule(lr){2-3} \cmidrule(lr){4-5} \cmidrule(lr){6-7} \cmidrule(lr){8-8} \cmidrule(lr){9-9} \cmidrule(lr){10-10}
& det. & loc. & det. & loc. & det. & loc. & det. & det. & det. & & \\
\midrule
PaDiM \textit{\fontsize{7}{10}\selectfont[ICPR'20]} & 79.0 & 94.4 & 50.8 & 90.9 & 75.9 & 91.4 & 91.8 & \underline{77.5} & 67.3 & 79.9  & 20 \\
CFlow \textit{\fontsize{7}{10}\selectfont[CVPR'20]}    & 74.8 & 93.8 & 50.8 & 92.4 & 75.0 & 93.8 & 85.4 & 71.5 & 55.7 & 77.0 & 15 \\
RD4AD \textit{\fontsize{7}{10}\selectfont[CVPR'22]}     & 89.5 & 96.5 & 60.4 & 96.0  & 87.8 & \underline{96.2}  & 97.3 & 67.6 & 66.8 & 84.2 & 20 \\
PatchCore \textit{\fontsize{7}{10}\selectfont[CVPR'22]} & \textbf{91.7} & \underline{97.0} & 60.3 & 96.4 & \underline{91.6} & \textbf{96.5} & \underline{98.6} & 76.1 & \textbf{69.3} & \underline{86.4} & 20 \\
SimpleNet \textit{\fontsize{7}{10}\selectfont[CVPR'23]} & 82.5 & 94.8 & \textbf{72.3} & \underline{97.5} & 76.2 & 77.1 & 94.7 & 69.1 & 62.4 & 80.7 & 10 \\
\midrule
\textbf{InvAD (Ours)} & \underline{90.3} & \textbf{97.8} & \underline{60.6} & \textbf{97.8} & \textbf{93.6} & \textbf{96.5} & \textbf{99.0} & \textbf{82.7} & \underline{67.7} & \textbf{87.2} & \textbf{88} \\
\bottomrule
\label{tab:BMAD_results}
\end{tabular}
\end{table*}

\subsection{Evaluation Metrics}
\label{sec:metrics}
Following \cite{omiad}, we employ Area Under the Receiver Operating Characteristic Curve (AU-ROC), Average Precision (AP), F1-score max (F1\_max) for both image- and pixel-level AD evaluation. We also adopt the Area Under the Per-Region-Overlap (AU-PRO) \cite{MVTecAD} metric to examine pixel-level anomaly localization.  
The Frames Per Second (FPS) metric is further applied to study the efficiency of inference.

\subsection{Implementations}
\label{sec:implementations}
To assess the effectiveness of the proposed method, we begin by training a Diffusion Transformer (DiT)-based diffusion model \cite{DiT} from scratch on each dataset with \Cref{eq:diffusion_objective2}, where we set \(\gamma = \mathbf{1}\) as DDPM \cite{DDPM}. To ensure fair comparisons, we employ the pre-trained EfficientNet-B4 \cite{EfficientNet} model on ImageNet-1k \cite{ImageNet} to encode the images into feature space as OmiAD \cite{omiad} and HVQ-Trans \cite{HVQ-Trans}. We train the diffusion model for 300 epochs across all datasets using the AdamW optimizer \cite{AdamW}. The total diffusion step \(T\) is set to 1000 during training using a linear noise schedule. For all experiments, we uniformly set \(S = 3\) to skip intermediate steps within the original \(T\) steps, and create the subset \(\tau_3 = [333, 666, 999]\) to acclerate inversion compuatation. We conduct all throughput measurements on an NVIDIA RTX 4090 (24GB) with batch size set to 64. See \textit{Supp. Mat.} for more details.

\begin{table}[t]
  \centering
  \caption{\textbf{Comparison of reconstruction-based approach (Recon.) against our inversion-based AD paradigm}, under different total diffusion steps $S$ and perturbation ratios $r$, in multi-class MVTecAD with image-level AU-ROC. }
  \label{tab:ablation_step}
  \resizebox{\columnwidth}{!}{%
    \begin{tabular}{llcccccc}
      \toprule
      \multicolumn{2}{c}{\textbf{Ratio} $r$} 
        & \multicolumn{6}{c}{\textbf{Total Diffusion Steps} $S$} \\
      \cmidrule(lr){3-8}
      \multicolumn{2}{c}{} 
        & \textbf{3} & \textbf{5} & \textbf{10} 
        & \textbf{50} & \textbf{100} & \textbf{1000} \\
      \midrule
\multirow{5}{*}{\rotatebox[origin=c]{90}{\textbf{Recon.}}}
        & 10\,\% & \na & \na & 66.6 & 97.7 & 97.9 & 98.0 \\
        & 20\,\% & \na & 64.7 & 97.9 & 91.3 & 98.2 & 98.2 \\
        & 40\,\% & 64.9 & 75.0 & 89.4 & 98.0 & 98.2 & 98.2 \\
        & 60\,\% & 64.9 & 68.4 & 87.1 & 97.3 & 97.8 & 98.1 \\
        & 80\,\% & 67.8 & 67.8 & 74.1 & 91.8 & 92.1 & 93.9 \\
      \midrule
      \multicolumn{2}{l}{\textbf{Inversion (Ours)}} & \textbf{99.0} & 98.9 & 98.4 & 96.0 & 95.7 & 95.4 \\
      \bottomrule
    \end{tabular}%
  }
\end{table}

\subsection{Comparison with SOTA AD Methods}
\label{sec:exp_performance}
We compare the quantitative results against prior multi-class AD competitors, including \textit{(i) Transformer-based methods} (HVQ-Trans \cite{HVQ-Trans}, UniAD \cite{uniad}), \textit{(ii) reconstruction-based methods} (RD4AD \cite{rd4ad}, DeSTSeg \cite{destseg}, SimpleNet \cite{simplenet}), \textit{(iii) diffusion-based methods} (DiAD \cite{DiAD}, OmiAD \cite{omiad}, MDM \cite{omiad}). The results are summarized in \Cref{tab:ad_visa_main}. It can be seen that our method generally yields the best image-level and highly competitive pixel-level detection performance on three datasets. Notably, our method consistently outperforms all the compared methods regarding inference efficiency by a large margin (\textit{i.e.}, FPS: 88.1 \textit{v.s.} 39.4 on MVTecAD). This can be attributed to the property of our proposed inversion-based AD paradigm, which enables high detection accuracy with fewer NFEs for acceleration. By contrast, conventional diffusion-based methods, such as DiAD \cite{DiAD}, require VAE decoding for reconstruction in the pixel space, thus resulting in slower inference. It is noteworthy that our method is also faster than OmiAD whose NFE is 1, which is primarily due to the heavy diffusion backbone and its modeling strategy in high-resolution feature space.

The results against diffusion-based AD models on single-class MVTecAD are displayed in \Cref{fig:scatter_plot_perf}. 
Similarly to the multi-class setting, InvAD yields substantially reduced inference latency with state-of-the-art performance, indicating the effectiveness of inversion modeling for AD. 

We next evaluate on BMAD to study the effectiveness of our InvAD on the medical domain. Here, we compare against PaDiM \cite{padim}, CFlow \cite{cflow}, RD4AD \cite{rd4ad}, PatchCore \cite{PatchCore}, and SimpleNet \cite{simplenet}.  We report image-level AU-ROC (det.), and pixel-level AU-ROC (loc.), if the ground-truth anomaly maps are provided. 
The results are presented in \Cref{tab:BMAD_results}. It can be confirmed that InvAD achieves state-of-the-art performance across diverse BMAD medical datasets with a substantially improved inference efficiency. This further evidences the capacity and robustness of our method in handling medical AD, other than industrial inspection.

\begin{table}[t]
\centering
\caption{\textbf{Generalizability evaluation} of our method when incorporated into other diffusion-based AD approaches on MVTecAD with image- and pixel-level AU-ROC, number of total parameters (\#Params) and FPS.}
\small
\setlength{\tabcolsep}{4pt}
\begin{tabular}{lcccc}
\toprule
\textbf{Method} & \multicolumn{2}{c}{\textbf{MVTecAD}} & \multicolumn{1}{c}{\textbf{FPS}} \\
\cmidrule(lr){2-3} 
& det. & loc. &  \\
\midrule
DiAD \textit{\fontsize{6.5}{8}\selectfont[AAAI'24]} & 97.2 & 96.8 & 0.1 \\
\rowcolor{OursBlue}
DiAD + \textbf{InvAD} & \textbf{98.2} {\scriptsize(+1.0)} & \textbf{97.5}{\scriptsize(+0.7)} & \textbf{88.1}{\scriptsize(+88)} \\
\midrule
MDM \textit{\fontsize{6.5}{8}\selectfont[ICML'25]} & 91.9 & 94.8 & 2.2 \\
\rowcolor{OursBlue}
MDM + \textbf{InvAD} & \textbf{98.2} {\scriptsize(+6.3)} & \textbf{97.5} {\scriptsize(+2.7)} & \textbf{63} {\scriptsize(+60.8)} \\
\bottomrule
\end{tabular}
\label{tab:generalization}
\end{table}

\subsection{Is Inversion Effective?}
\label{sec:analy_inversion}
Our approach builds on the inversion modeling to realize reconstruction-free AD.
%, rather than following a reconstructive manner.
To investigate the validity of this design, we vary two key parameters in reconstruction-based methods, total diffusion steps $S$ and perturbation timesteps $t$, to gain a deeper insight into the effect. In particular, instead of inspecting $t$ directly, we study the ratio of $r = t/S$ as it better captures the perturbation strength across different $S$. The results are summarized in \Cref{tab:ablation_step}, where we only modify the inference stage to examine the accuracy (\textit{i.e.,} image-level AU-ROC) change. It can be confirmed that for reconstruction-based methods, the parameter tuning efforts can be highly burdensome to yield ideal performance. While a larger $S$ tends to perform better, it induces computationally expensive denoising iterations. 

By contrast, our inversion-based method can achieve the best performance even with an extremely small inversion step setting (\textit{e.g.}, $S=3, 5$ in the bottom row of \Cref{tab:ablation_step}), with no reliance on perturbation timesteps as it is not involved in our modeling. This further evidences the superiority of our InvAD for the task of image AD. 

\subsection{Generalizability Evaluation}
\label{sec:plug_and_play}
As our InvAD is designed for the inference stage, it can be incorporated into existing diffusion-based AD models in a plug-and-play manner. To evaluate the effectiveness, we incorporate InvAD into the inference stage for DiAD \cite{DiAD} and MDM \cite{omiad}, and report the image- and pixel-level AU-ROC and the FPS metrics on multi-class MVTecAD. It can be seen from  \Cref{tab:generalization} that InvAD leads to an accuracy gain for both methods with substantial inference acceleration. 
In particular, since MDM learns to optimize the noise residual during training, it suits better for our method, as reflected in the remarkable improvement of accuracy.
This suggests that InvAD is model-agnostic and can be inserted as a plug-and-play module into different diffusion-based AD techniques.

\begin{table}[t]
  \centering
  \small
  \caption{\textbf{Component ablation} on multi-class MVTecAD. FDM denotes feature space diffusion models. S-Inv and M-Inv denote single- and multi-step inversion, resepectively.}
  \label{tab:ablation}
  \setlength{\tabcolsep}{7pt}
  \begin{tabular}{@{}lcccc@{}}
    \toprule
    Setting & \textbf{FDM} & \textbf{S-Inv} & \textbf{M-Inv} & \textbf{mAD} \\
    \midrule
    A1 & \Yes &        &        & 57.3       \\
    A2 &      & \Yes   &        & 44.9     \\
    A3 & \Yes & \Yes   &        & 71.0      \\
    \rowcolor{OursBlue}
    A4 (\textbf{Ours})  & \Yes &        & \Yes      & \textbf{83.7}  \\
    \bottomrule
  \end{tabular}
\end{table}

\subsection{Component Ablation}
\label{sec:component_ablation}
Our framework consists of two components: (i) feature space diffusion model (FDM), and (ii) multi-step inversion. To individually validate the contribution, we evaluate the AD performance of each component as well as their combinations on the multi-class MVTecAD. \Cref{tab:ablation} reports the results of four scenarios (A1-A4). In the case of A1, we skip the inversion (lines 3-8 of \Cref{alg:proposed_method}), then calculate the anomaly score as in \cite{mahalnobisAD}. This leads to a significantly degredation in AD performance compared with our complete framework (A4), highlighting the importance of inversion. Furthermore, for the FDM-free case of A2  (\textit{i.e.,} pixel-space diffusion), single-step inversion also degrades accuracy. Combining FDM and single-step inversion (A3) substantially improves AD accuracy, and multi-step inversion (A4) brings further accuracy gains.

\begin{table}[t]
\centering
\small
\setlength{\tabcolsep}{7pt} 
\caption{\textbf{Design ablation of feature space encoding modules} on multi-class MVTecAD with image- and pixel-level AU-ROC, number of total parameters (\#Params), and FPS.}
\begin{tabular}{lcccc}
\toprule
\textbf{Model} & \multicolumn{2}{c}{\textbf{MVTecAD}}  & \textbf{\# Params} & \textbf{FPS} \\
\cmidrule(lr){2-3} &
det. & loc. & & \\
\midrule
% VAE & 92.4 & 94.2 & 1289M & 52.6 \\
ViT-B \cite{ViT} & 94.2 & 95.5 & 1312M & 85.3 \\
DINO-B \cite{dino} & 96.1 & \textbf{97.7} & 1312M & 85.3 \\
EfficientNet-B4 \cite{EfficientNet} & \textbf{99.0} & 96.9 & \textbf{1243M} & \textbf{88.1} \\
\bottomrule
\end{tabular}
\label{tab:backbone_auc}
\end{table}

\begin{table}[t]
\centering
\small
\setlength{\tabcolsep}{5pt} 
\caption{\textbf{Design ablation of diffusion architectures} on multi-class MVTecAD with image- and pixel-level AU-ROC, and FPS.}
\begin{tabular}{lccccc}
\toprule
\textbf{Model} & \multicolumn{2}{c}{\textbf{MVTecAD}}  & \textbf{\# Params} & \textbf{FPS} \\
\cmidrule(lr){2-3} 
& det. & loc. & & \\
\midrule
MLP \cite{mar} & 97.2 & 96.8 & 351M & 241 \\
UNet \cite{DDPM} & 98.0 & 97.3 & 1433M & \textbf{288} \\
DiT-base \cite{DiT} & 93.8 & 95.9 & \textbf{155M} & 240 \\
DiT-gigant \cite{DiT} & \textbf{99.0} & \textbf{97.5} & 1223M & 88 \\
\bottomrule
\end{tabular}
\label{tab:model_comparison}
\end{table}

\begin{table}[t]
  \centering
  \small
  \caption{\textbf{Scoring scheme ablation} under different total diffusion steps $S$ in multi-class MVTecAD with image-level AU-ROC.}
  \label{tab:ablation_scoring}
  \resizebox{\columnwidth}{!}{%
    \begin{tabular}{lcccccc}
      \toprule
      \textbf{Total Diffusion Steps} $S$
        & \textbf{3} & \textbf{5} & \textbf{10} 
        & \textbf{50} & \textbf{100} & \textbf{1000} \\
      \midrule
    \textbf{NLL} & 96.1 & 95.2 & 93.0 & 89.7 & 89.4 & 89.1 \\
      \textbf{Diff} & 98.4 & 98.1 & 95.8 & 83.2 & 82.2 & 82.0 \\
      \rowcolor{OursBlue}
      \textbf{NLL + Diff (Ours)} & \textbf{99.0} & \textbf{98.9} & \textbf{98.4} & \textbf{96.0} & \textbf{95.7} & \textbf{95.4} \\
      \bottomrule
    \end{tabular}%
  }
\end{table}

\subsection{Backbone Design Ablation}
Our inversion-based method is considered in the pre-trained feature space rather than pixel space. In general, the selection of feature extractors heavily impacts the performance of AD methods \cite{adpretrain,DiAD,DDAD}. In \Cref{tab:backbone_auc}, we evaluate the performance with different backbones including ViT-B \cite{ViT}, DINO-B \cite{dino}, and EfficientNet-B4 \cite{EfficientNet}. As expected, stronger backbones lead to better performance. It should be noted that EfficientNet is a common backbone in diffusion-based AD \cite{decodiff, omiad, vpot}. We follow this standard choice for fair comparisons, while contributing a novel inversion-based paradigm to jointly harvest accuracy and efficiency. 

Also, we study the influence of the design of diffusion modeling backbones on AD performance in \Cref{tab:model_comparison}. In particular, we compare four types of backbones: MLP, UNet, DiT-base, and DiT-gigant \cite{DiT}. In contrast to the feature encoding modules, we observe that the performance is generally similar, even with the simple MLP implementation. 
This suggests that our method is robust to diffusion architectures, but also relies on the quality of embeddings derived with feature extractors, as prior methods \cite{omiad,PatchCore}. 
See \textit{Supp. Mat.} for a detailed description of backbones.

\subsection{Scoring Scheme Ablation}
\label{sec:reverse_scoring}
As explained in \Cref{sec:anomaly_score}, our anomaly scoring scheme considers both negative-log-likelihood (NLL) in the noise prior and the spatial difference of Euclidean norm (Diff) to alleviate the \textit{reverse-scoring} problem \cite{DoGMKnow,OODTP}. To study the advantage of this strategy, we compare the performance in \Cref{tab:ablation_scoring} with the conventional NLL-based scoring on MVTecAD. It can be seen that our scoring scheme (\textit{i.e.}, NLL + Diff) contributes to robustness to the diffusion step setting $S$, while NLL or Diff solely do not achieve this. This demonstrates that our scoring scheme plays a key role in the proposed InvAD paradigm to realize reconstruction-free AD, where $S$ can be set to be very small for efficiency.

\label{sec:analy_diffusion_arch}

\begin{figure}[t]
  \centering
  \vspace{-8pt}
  \includegraphics[width=1.0\linewidth]{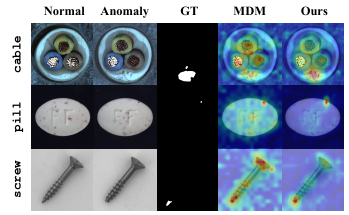}
  \caption{\textbf{Visualization of anomaly localization} against MDM \cite{omiad} on MVTecAD. GT displays the ground-truth anomaly map.}
  \label{fig:visualization}
  \vspace{-12pt}
\end{figure}

\subsection{Visualization Analysis}
\label{sec:visualization}
We qualitatively evaluate in \Cref{fig:visualization} the pixel-level AD performance of InvAD against MDM \cite{omiad}. Similar to the quantitative evaluation in \Cref{tab:ad_visa_main}, InvAD generally achieves more precise anomaly localization with fewer false positives than MDM, which again, demonstrates that our method improves efficiency and accuracy while preserving interpretability for anomaly locations without reconstruction. 
\section{Conclusion}
\label{sec:conclusion}
We propose a simple yet effective diffusion-based AD paradigm, “\textit{detection via noising in latent space}”, to address the fundamental challenges within the conventional “\textit{detection via denoising in RGB space}”.
Our method,  \textbf{InvAD}, is reconstruction-free and directly infers the final latent state with the DDIM inversion mechanism. To pursue efficiency, we simply enforce very few inversion steps to noise a test sample, yet manage to maintain high detection accuracy as the typicality can still be accurately scored. Due to the reconstruction-free design, InvAD is inherently tuning-free for the perturbation timestep. We experimentally demonstrate that InvAD achieves state-of-the-art detection performance, especially regarding inference efficiency.

\noindent\textbf{Limitation and Future Work.} While our method can achieve satisfactory performance, it still needs more than one NFE. To further improve efficiency (\textit{i.e.}, 1 NFE), we will consider introducing diffusion distillation to compactly compress the progressive inversion process into only one single step. Also, exploring improved or task-specific inversion mechanism is an interesting future direction.

%Speed, simplicity, Accuracy%
\section{Acknowledgment}
\label{sec:acknowledgment}

This work was supported by JSPS KAKENHI Grant Number JP25H01110 and the TAKEUCHI research grant.
\bibliographystyle{unsrt} 
{
\small
\bibliography{reference/ref}

@String(CVPR= {IEEE Conf. Comput. Vis. Pattern Recog.})

@String(ICCV= {Int. Conf. Comput. Vis.})

@String(ECCV= {Eur. Conf. Comput. Vis.})

@String(NIPS= {Adv. Neural Inform. Process. Syst.})

@String(ICPR = {Int. Conf. Pattern Recog.})

@String(ICLR = {Int. Conf. Learn. Represent.})

@String(AAAI = {AAAI})

@String(CVPRW= {IEEE Conf. Comput. Vis. Pattern Recog. Worksh.})

@String(CVPR  = {CVPR})

@String(ICCV  = {ICCV})

@String(ECCV  = {ECCV})

@String(NIPS  = {NeurIPS})

@String(ICPR  = {ICPR})

@String(ICLR  = {ICLR})

@String(CVPRW= {CVPRW})

@article{Yang2020,
  author    = {Yang J and Li S and Wang Z and Dong H and Wang J and Tang S},
  title     = {Using Deep Learning to Detect Defects in Manufacturing: A Comprehensive Survey and Current Challenges},
  journal   = {Materials (Basel)},
  volume    = {13},
  number    = {24},
  pages     = {5755},
  year      = {2020},
  month     = {Dec 16},
  doi       = {10.3390/ma13245755},
  pmid      = {33339413},
  pmcid     = {PMC7766692}
}

@inproceedings{
bercea2023generalizing,
title={Generalizing Unsupervised Anomaly Detection: Towards Unbiased Pathology Screening},
author={Cosmin I. Bercea and Benedikt Wiestler and Daniel Rueckert and Julia A Schnabel},
booktitle={Medical Imaging with Deep Learning},
year={2023},
url={https://openreview.net/forum?id=8ojx-Ld3yjR}
}

@inproceedings{DDPM,
    author = {Ho, Jonathan and Jain, Ajay and Abbeel, Pieter},
    title = {Denoising diffusion probabilistic models},
    year = {2020},
    booktitle = {NeurIPS},
}

@inproceedings{DiffusionModel,
    author = {Sohl-Dickstein, Jascha and Weiss, Eric A. and Maheswaranathan, Niru and Ganguli, Surya},
    title = {Deep unsupervised learning using nonequilibrium thermodynamics},
    year = {2015},
    booktitle = {ICML},
}

@article{DiAD,
  title = {A Diffusion-Based Framework for Multi-Class Anomaly Detection},
  author = {He, Haoyang and Zhang, Jiangning and Chen, Hongxu and Chen, Xuhai and Li, Zhishan and Chen, Xu and Wang, Yabiao and Wang, Chengjie and Xie, Lei},
  journal = {Proceedings of the AAAI Conference on Artificial Intelligence},
  volume = {38},
  number = {8},
  pages = {8472--8480},
  year = {2024},
  month = {Mar.},
  doi = {10.1609/aaai.v38i8.28690},
  url = {https://ojs.aaai.org/index.php/AAAI/article/view/28690},
}

@inproceedings{RemoveAsNoise,
	author = {Lu, Fanbin and Yao, Xufeng and Fu, Chi-Wing and Jia, Jiaya},
	year = {2023},
	month = {10},
	pages = {16120-16129},
	title = {Removing Anomalies as Noises for Industrial Defect Localization},
	doi = {10.1109/ICCV51070.2023.01481}
}

@inproceedings{vpot,
title={Vague Prototype-Oriented Diffusion Model for Multi-Class Anomaly Detection},
author={Yuxin Li and Yaoxuan Feng and Bo Chen and Wenchao Chen and Yubiao Wang and Xinyue Hu and Baolin Sun and Chunhui Qu and Mingyuan Zhou},
booktitle={Forty-first International Conference on Machine Learning},
year={2024},
}

@INPROCEEDINGS{DiffAD,
  author={Zhang, Xinyi and Li, Naiqi and Li, Jiawei and Dai, Tao and Jiang, Yong and Xia, Shu-Tao},
  booktitle={2023 IEEE/CVF International Conference on Computer Vision (ICCV)}, 
  title={Unsupervised Surface Anomaly Detection with Diffusion Probabilistic Model}, 
  year={2023},
  volume={},
  number={},
  pages={6759-6768},
  doi={10.1109/ICCV51070.2023.00624}}

@article{GLAD,
  title={GLAD: Towards Better Reconstruction with Global and Local Adaptive Diffusion Models for Unsupervised Anomaly Detection},
  author={Yao, Hang and Liu, Ming and Wang, Haolin and Yin, Zhicun and Yan, Zifei and Hong, Xiaopeng and Zuo, Wangmeng},
  journal={arXiv preprint arXiv:2406.07487},
  year={2024}
}

@article{DiffusionAD,
  title={DiffusionAD: Norm-guided One-step Denoising Diffusion for Anomaly Detection},
  author={Zhang, Hui and Wang, Zheng and Wu, Zuxuan and Jiang, Yu-Gang},
  journal={arXiv preprint arXiv:2303.08730},
  year={2023}
}

@article{DDAD,
  title={Anomaly Detection with Conditioned Denoising Diffusion Models},
  author={Mousakhan, Arian and Brox, Thomas and Tayyub, Jawad},
  journal={arXiv preprint arXiv:2305.15956},
  year={2023}
}

@article{EfficientAD,
  title={EfficientAD: Accurate Visual Anomaly Detection at Millisecond-Level Latencies},
  author={Kilian Batzner and Lars Heckler and Rebecca K{\"o}nig},
  journal= {WACV},
  year={2023},
}

@article{PatchCore,
  title={Towards Total Recall in Industrial Anomaly Detection},
  author={Karsten Roth and Latha Pemula and Joaquin Zepeda and Bernhard Scholkopf and Thomas Brox and Peter Gehler},
  journal= {CVPR},
  year={2021},
}

@article{FastFlow,
author = {Um, Taegeon and Oh, Byungsoo and Seo, Byeongchan and Kweun, Minhyeok and Kim, Goeun and Lee, Woo-Yeon},
title = {FastFlow: Accelerating Deep Learning Model Training with Smart Offloading of Input Data Pipeline},
year = {2023},
issue_date = {January 2023},
publisher = {VLDB Endowment},
volume = {16},
number = {5},
journal = {Proc. VLDB Endow.},
month = jan,
pages = {1086–1099},
numpages = {14}
}

@inproceedings{HVQ-Trans,
  title={Hierarchical Vector Quantized Transformer for Multi-class Unsupervised Anomaly Detection},
  author={Ruiying Lu and YuJie Wu and Long Tian and Dongsheng Wang and Bo Chen and Xiyang Liu and Ruimin Hu},
  booktitle= NeurIPS,
  year={2023},
  month={December},
  article_no={370},
  pages={8487--8500},
  doi = "10.48550/arXiv.2310.14228",
}

@inproceedings{ViSA,
  title={SPot-the-Difference Self-Supervised Pre-training for Anomaly Detection and Segmentation},
  author={Yang Zou and Jongheon Jeong and Latha Pemula and Dongqing Zhang and Onkar Dabeer},
  booktitle= ECCV,
  year={2022},
  doi = "10.1007/978-3-031-20056-4_23"
}

@article{MVTecAD,
  title={MVTec AD — A Comprehensive Real-World Dataset for Unsupervised Anomaly Detection},
  author={Paul Bergmann and Michael Fauser and David Sattlegger and Carsten Steger},
  journal= CVPR,
  year={2019},
  pages={9584-9592},
  doi = "10.1109/CVPR.2019.00982",
}

@inproceedings{DDIM,
  author       = {Jiaming Song and
                  Chenlin Meng and
                  Stefano Ermon},
  title        = {Denoising Diffusion Implicit Models},
  booktitle    = {9th International Conference on Learning Representations, {ICLR} 2021,
                  Virtual Event, Austria, May 3-7, 2021},
  publisher    = {OpenReview.net},
  year         = {2021},
  url          = {https://openreview.net/forum?id=St1giarCHLP},
  bibsource    = {dblp computer science bibliography, https://dblp.org}
}

@inproceedings{AdamW,
  title={Decoupled Weight Decay Regularization},
  author={Ilya Loshchilov and Frank Hutter},
  booktitle= ICLR,
  year={2017},
  doi = "10.48550/arXiv.1711.05101"
}

@article{EfficientNet,
  title={EfficientNet: Rethinking Model Scaling for Convolutional Neural Networks},
  author={Tan, Mingxing and Le, Quoc V},
  booktitle={Proceedings of the 36th International Conference on Machine Learning},
  pages={6105--6114},
  year={2019},
  organization= ICML,
  doi = "10.48550/arXiv.1905.11946"
}

@article{ViT,
  title={An Image is Worth 16x16 Words: Transformers for Image Recognition at Scale},
  author={Dosovitskiy, Alexey and Beyer, Lucas and Kolesnikov, Alexander and Weissenborn, Dirk and Zhai, Xiaohua and Unterthiner, Thomas and Dehghani, Mostafa and Minderer, Matthias and Heigold, Georg and Gelly, Sylvain and Uszkoreit, Jakob and Houlsby, Neil},
  journal={Computer Vision and Image Understanding},
  year={2020},
  volume={abs/2010.11929},
  doi = "10.48550/arXiv.2010.11929",
}

@inproceedings{ImageNet,
  author={Deng, Jia and Dong, Wei and Socher, Richard and Li, Li-Jia and Kai Li and Li Fei-Fei},
  booktitle={2009 IEEE Conference on Computer Vision and Pattern Recognition}, 
  title={ImageNet: A large-scale hierarchical image database}, 
  year={2009},
  volume={},
  number={},
  pages={248-255},
  doi={10.1109/CVPR.2009.5206848}
}

@InProceedings{UNet,
    author="Ronneberger, Olaf
    and Fischer, Philipp
    and Brox, Thomas",
    title="U-Net: Convolutional Networks for Biomedical Image Segmentation",
    booktitle="Medical Image Computing and Computer-Assisted Intervention -- MICCAI 2015",
    year="2015",
    publisher="Springer International Publishing",
    address="Cham",
    pages="234--241",
    isbn="978-3-319-24574-4"
}

@InProceedings{TransFusion,
    author="Fu{\v{c}}ka, Matic
    and Zavrtanik, Vitjan
    and Sko{\v{c}}aj, Danijel",
    title="TransFusion -- A Transparency-Based Diffusion Model for Anomaly Detection",
    booktitle="Computer Vision -- ECCV 2024",
    year="2025",
    publisher="Springer Nature Switzerland",
    address="Cham",
    pages="91--108",
    isbn="978-3-031-72761-0"
}

@misc{
ScoreDD,
title={Unsupervised Visual Anomaly Detection with Score-Based Generative Model},
author={Yapeng Teng and HaoYang LI and Fuzhen Cai and Ming Shao and Siyu Xia},
year={2023},
url={https://openreview.net/forum?id=j6sAOkvn4GI}
}

@inproceedings{SBGM,
  title={Score-Based Generative Modeling through Stochastic Differential Equations},
  author={Yang Song and Jascha Sohl-Dickstein and Diederik P Kingma and Abhishek Kumar and Stefano Ermon and Ben Poole},
  booktitle={International Conference on Learning Representations},
  year={2021},
  url={https://openreview.net/forum?id=PxTIG12RRHS}
}

@inproceedings{AnoDDPM,
  author={Wyatt, Julian and Leach, Adam and Schmon, Sebastian M. and Willcocks, Chris G.},
  booktitle={2022 IEEE/CVF Conference on Computer Vision and Pattern Recognition Workshops (CVPRW)}, 
  title={AnoDDPM: Anomaly Detection with Denoising Diffusion Probabilistic Models using Simplex Noise}, 
  year={2022},
  volume={},
  number={},
  pages={649-655},
  doi={10.1109/CVPRW56347.2022.00080}
}

@misc{maediff,
      title={MAEDiff: Masked Autoencoder-enhanced Diffusion Models for Unsupervised Anomaly Detection in Brain Images}, 
      author={Rui Xu and Yunke Wang and Bo Du},
      year={2024},
      eprint={2401.10561},
      archivePrefix={arXiv},
      primaryClass={eess.IV},
      url={https://arxiv.org/abs/2401.10561}, 
}

@inproceedings{DoGMKnow,
  author       = {Eric T. Nalisnick and
                  Akihiro Matsukawa and
                  Yee Whye Teh and
                  Dilan G{\"{o}}r{\"{u}}r and
                  Balaji Lakshminarayanan},
  title        = {Do Deep Generative Models Know What They Don't Know?},
  booktitle    = {7th International Conference on Learning Representations, {ICLR} 2019,
                  New Orleans, LA, USA, May 6-9, 2019},
  publisher    = {OpenReview.net},
  year         = {2019},
  url          = {https://openreview.net/forum?id=H1xwNhCcYm},
  bibsource    = {dblp computer science bibliography, https://dblp.org}
}

@inproceedings{OODTP,
  author    = {Genki Osada and Takahashi Tsubasa and Budrul Ahsan and Takashi Nishide},
  title     = {Out-of-Distribution Detection with Reconstruction Error and Typicality-based Penalty},
  booktitle = {Proceedings of the IEEE/CVF Winter Conference on Applications of Computer Vision (WACV)},
  year      = {2023},
  month     = {January},
  pages     = {},
  publisher = {IEEE},
  address   = {},
}

@inproceedings{omiad,
  author    = {Yaoxuan Feng and Wenchao Chen and Yuxin Li and Bo Chen and Yubiao Wang and Zixuan Zhao and Hongwei Liu and Mingyuan Zhou},
  title     = {{OmiAD: One-Step Adaptive Masked Diffusion Model for Multi-class Anomaly Detection via Adversarial Distillation}},
  booktitle = {Proceedings of the International Conference on Machine Learning (ICML)},
  year      = {2025},
  address   = {Vancouver, Canada},
  note      = {To appear},
}

@inproceedings{decodiff,
  title={Correcting deviations from normality: A reformulated diffusion model for multi-class unsupervised anomaly detection},
  author={Beizaee, Farzad and Lodygensky, Gregory A and Desrosiers, Christian and Dolz, Jose},
  booktitle={Proceedings of the Computer Vision and Pattern Recognition Conference},
  pages={19088--19097},
  year={2025}
}

@inproceedings{DiT,
  author={Peebles, William and Xie, Saining},
  booktitle={2023 IEEE/CVF International Conference on Computer Vision (ICCV)}, 
  title={Scalable Diffusion Models with Transformers}, 
  year={2023},
  volume={},
  number={},
  pages={4172-4182},
  doi={10.1109/ICCV51070.2023.00387}
}

@inproceedings{MPDD,
  author={Jezek, Stepan and Jonak, Martin and Burget, Radim and Dvorak, Pavel and Skotak, Milos},
  booktitle={2021 13th International Congress on Ultra Modern Telecommunications and Control Systems and Workshops (ICUMT)}, 
  title={Deep learning-based defect detection of metal parts: evaluating current methods in complex conditions}, 
  year={2021},
  volume={},
  number={},
  pages={66-71},
  doi={10.1109/ICUMT54235.2021.9631567}
}

@inproceedings{diffpath,
 author = {Heng, Alvin and Thiery, Alexandre H. and Soh, Harold},
 booktitle = {Advances in Neural Information Processing Systems},
 editor = {A. Globerson and L. Mackey and D. Belgrave and A. Fan and U. Paquet and J. Tomczak and C. Zhang},
 pages = {43952--43974},
 publisher = {Curran Associates, Inc.},
 title = {Out-of-Distribution Detection with a Single Unconditional Diffusion Model},
 url = {https://proceedings.neurips.cc/paper_files/paper/2024/file/4dc37a7bc61057252ce043fa3b83aac2-Paper-Conference.pdf},
 volume = {37},
 year = {2024}
}

@inproceedings{uniad,
author = {You, Zhiyuan and Cui, Lei and Shen, Yujun and Yang, Kai and Lu, Xin and Zheng, Yu and Le, Xinyi},
title = {A unified model for multi-class anomaly detection},
year = {2022},
isbn = {9781713871088},
publisher = {Curran Associates Inc.},
address = {Red Hook, NY, USA},
booktitle = {Proceedings of the 36th International Conference on Neural Information Processing Systems},
articleno = {330},
numpages = {14},
location = {New Orleans, LA, USA},
series = {NIPS '22}
}

@inproceedings{destseg,
  title={DeSTSeg: Segmentation Guided Denoising Student-Teacher for Anomaly Detection},
  author={Zhang, Xuan and Li, Shiyu and Li, Xi and Huang, Ping and Shan, Jiulong and Chen, Ting},
  booktitle={Proceedings of the IEEE/CVF Conference on Computer Vision and Pattern Recognition},
  pages={3914--3923},
  year={2023}
}

@inproceedings{rd4ad,
  author={Deng, Hanqiu and Li, Xingyu},
  booktitle={2022 IEEE/CVF Conference on Computer Vision and Pattern Recognition (CVPR)}, 
  title={Anomaly Detection via Reverse Distillation from One-Class Embedding}, 
  year={2022},
  volume={},
  number={},
  pages={9727-9736},
  doi={10.1109/CVPR52688.2022.00951}
}

@inproceedings{mar,
author = {Li, Tianhong and Tian, Yonglong and Li, He and Deng, Mingyang and He, Kaiming},
title = {Autoregressive image generation without vector quantization},
year = {2025},
isbn = {9798331314385},
publisher = {Curran Associates Inc.},
address = {Red Hook, NY, USA},
booktitle = {Proceedings of the 38th International Conference on Neural Information Processing Systems},
articleno = {1797},
numpages = {22},
location = {Vancouver, BC, Canada},
series = {NIPS '24}
}

@inproceedings{dino,
  author={Caron, Mathilde and Touvron, Hugo and Misra, Ishan and Jegou, Hervé and Mairal, Julien and Bojanowski, Piotr and Joulin, Armand},
  booktitle={2021 IEEE/CVF International Conference on Computer Vision (ICCV)}, 
  title={Emerging Properties in Self-Supervised Vision Transformers}, 
  year={2021},
  volume={},
  number={},
  pages={9630-9640},
  doi={10.1109/ICCV48922.2021.00951}
}

@incollection{torch,
title = {PyTorch: An Imperative Style, High-Performance Deep Learning Library},
author = {Paszke, Adam and Gross, Sam and Massa, Francisco and Lerer, Adam and Bradbury, James and Chanan, Gregory and Killeen, Trevor and Lin, Zeming and Gimelshein, Natalia and Antiga, Luca and Desmaison, Alban and Kopf, Andreas and Yang, Edward and DeVito, Zachary and Raison, Martin and Tejani, Alykhan and Chilamkurthy, Sasank and Steiner, Benoit and Fang, Lu and Bai, Junjie and Chintala, Soumith},
booktitle = {Advances in Neural Information Processing Systems 32},
pages = {8024--8035},
year = {2019},
publisher = {Curran Associates, Inc.},
url = {http://papers.neurips.cc/paper/9015-pytorch-an-imperative-style-high-performance-deep-learning-library.pdf}
}

@inproceedings{cflow,
  author={Pumarola, Albert and Popov, Stefan and Moreno-Noguer, Francesc and Ferrari, Vittorio},
  booktitle={2020 IEEE/CVF Conference on Computer Vision and Pattern Recognition (CVPR)}, 
  title={C-Flow: Conditional Generative Flow Models for Images and 3D Point Clouds}, 
  year={2020},
  volume={},
  number={},
  pages={7946-7955},
  doi={10.1109/CVPR42600.2020.00797}
}

@inproceedings{bmad,
  author={Bao, Jinan and Sun, Hanshi and Deng, Hanqiu and He, Yinsheng and Zhang, Zhaoxiang and Li, Xingyu},
  booktitle={2024 IEEE/CVF Conference on Computer Vision and Pattern Recognition Workshops (CVPRW)}, 
  title={BMAD: Benchmarks for Medical Anomaly Detection}, 
  year={2024},
  volume={},
  number={},
  pages={4042-4053},
  doi={10.1109/CVPRW63382.2024.00408}
 }

@inproceedings{padim,
    author="Defard, Thomas
    and Setkov, Aleksandr
    and Loesch, Angelique
    and Audigier, Romaric",
    editor="Del Bimbo, Alberto
    and Cucchiara, Rita
    and Sclaroff, Stan
    and Farinella, Giovanni Maria
    and Mei, Tao
    and Bertini, Marco
    and Escalante, Hugo Jair
    and Vezzani, Roberto",
    title="PaDiM: A Patch Distribution Modeling Framework for Anomaly Detection and Localization",
    booktitle="Pattern Recognition. ICPR International Workshops and Challenges",
    year="2021",
    publisher="Springer International Publishing",
    address="Cham",
    pages="475--489",
    isbn="978-3-030-68799-1"
}

@String{CVPR  = {CVPR}}

@String{CVPRW = {CVPRW}}

@String{ICCV  = {ICCV}}

@String{ECCV  = {ECCV}}

@String{ICPR  = {ICPR}}

@String{NIPS  = {NeurIPS}}

@String{ICLR  = {ICLR}}

@String{AAAI  = {AAAI}}

@String{ICML   = {ICML}}

@String{WACV   = {WACV}}

@String{MICCAI = {MICCAI}}

@String{ICUMT  = {ICUMT}}

@String{IPMI   = {IPMI}}

@inproceedings{mahalnobisAD,
  author    = {Rippel, Oliver and Mertens, Patrick and Merhof, Dorit},
  title     = {Modeling the Distribution of Normal Data in Pre-Trained Deep Features for Anomaly Detection},
  booktitle = ICPR,
  pages     = {6726--6733},
  year      = {2021},
  doi       = {10.1109/ICPR48806.2021.9412109}
}

@inproceedings{simplenet,
  author    = {Liu, Zhikang and Zhou, Yiming and Xu, Yuansheng and Wang, Zilei},
  title     = {{SimpleNet}: A Simple Network for Image Anomaly Detection and Localization},
  booktitle = CVPR,
  pages     = {20402--20411},
  year      = {2023},
  doi       = {10.1109/CVPR52729.2023.01954},
  url       = {https://openaccess.thecvf.com/content/CVPR2023/papers/Liu_SimpleNet_A_Simple_Network_for_Image_Anomaly_Detection_and_Localization_CVPR_2023_paper.pdf}
}

@inproceedings{adpretrain,
  author    = {Yao, Xincheng and Luo, Yan and Qian, Zefeng and Zhang, Chongyang},
  title     = {{ADP}retrain: Advancing Industrial Anomaly Detection via Anomaly Representation Pretraining},
  booktitle = NEURIPS,
  year      = {2025},
  url       = {https://openreview.net/forum?id=mHfpziOtTW}
}

@inproceedings{mad,
  author    = {Beizaee, Farzad and Lodygensky, Gregory and Desrosiers, Christian and Dolz, Jose},
  title     = {{MAD-AD}: Masked Diffusion for Unsupervised Brain Anomaly Detection},
  booktitle = IPMI,
  pages     = {139--153},
  year      = {2025},
  publisher = {Springer}
}

@inproceedings{dte,
  author    = {Livernoche, Victor and Jain, Vineet and Hezaveh, Yashar and Ravanbakhsh, Siamak},
  title     = {On Diffusion Modeling for Anomaly Detection},
  booktitle = ICLR,
  year      = {2024},
  url       = {https://openreview.net/forum?id=lR3rk7ysXz}
}

@inproceedings{dhariwal2021diffusion,
  author    = {Dhariwal, Prafulla and Nichol, Alexander Quinn},
  title     = {Diffusion Models Beat {GAN}s on Image Synthesis},
  booktitle = NEURIPS,
  year      = {2021},
  url       = {https://openreview.net/forum?id=AAWuCvzaVt}
}
}
\clearpage
\setcounter{page}{1}
\maketitlesupplementary

% ----------- Supplementary Content Starts Here -----------
This document supplements our main paper, ``InvAD: Inversion-based Reconstruction-Free Anomaly Detection with
Diffusion Models''. The reference indices are consistent with those used in the main paper. 

\section{Dataset Details}
\label{sec:sup_dataset_details}
We utilize four publicly available anomaly detection (AD) benchmarks for evaluation: MVTecAD \cite{MVTecAD}, VisA \cite{ViSA}, MPDD \cite{MPDD}, and BMAD \cite{bmad}. \Cref{tab:sup_dataset_stats} summarizes the key statistics of the datasets used in our experiments, including the number of classes, domain, image resolution, and sample counts for training and testing. All of these datasets follow an unsupervised AD setting. In other words, the training set only consists of defect-free normal images, and the test set consists of both normal and anomalous images.

\section{Implementation Details}
\label{sec:sup_implementation_details}

Our implementation is available at: \url{https://github.com/SkyShunsuke/InversionAD}.

\subsection{Data Preparation}
We resize all training images to 256$\times$256 pixels and do not apply any data augmentation. Image normalization is performed using the ImageNet dataset’s channel-wise mean ([0.485, 0.456, 0.406]) and standard deviation ([0.229, 0.224, 0.225]).

\subsection{Model Setup}
We adopt the original DiT architecture \cite{DiT}
as the backbone of our diffusion modeling, but remove all class‐conditioning so that our method does not depend on class labels. This design allows DiT to be applied in scenarios where ground‐truth annotations are unavailable or impractical (\textit{e.g.}, inspection of diverse products on a manufacturing line). We set the number of transformer blocks and embedding dimensions to 16 and 2048 for DiT-gigant, and to 8 and 1024 for DiT-base, respectively. For the UNet‐based diffusion models evaluated in \Cref{tab:model_comparison} (in the main paper), we follow the architectural specifications of the original DDPM implementation \cite{DDPM}, consistent with prior work \cite{decodiff,DiAD}. For the MLP‐based diffusion models, we stack multiple MLP blocks and apply AdaLN-Zero conditioning \cite{DiT}, as in \cite{mar}. We summarize all major hyperparameters for diffusion modeling in \Cref{tab:sup_configurations}.

\subsection{Backbone Setup}
\label{sec:sup_backbone_details}
For fair comparisons with previous works \cite{omiad,HVQ-Trans}, we employ pre-trained EfficientNet-B4 \cite{EfficientNet} as our feature extractor. To capture anomalies at different scales, we first extract feature maps from the 1st to the 4th blocks from EfficientNet-B4, and then apply bilinear interpolation followed by channel-wise concatenation, creating $\mathbf{z} \in \mathbb{R}^{272 \times 16 \times 16}$, as in \cite{omiad,HVQ-Trans}. 

We evaluate in \Cref{tab:backbone_auc} in the main paper the robustness of our method to the different feature space encoding modules: ViT-B \cite{ViT} and DINO-base \cite{dino}. 
Both modules employ the vision transformer-based architecture and use ImageNet-1k as the pre-training dataset. The only difference comes from the pre-training scheme (\ie, supervised for ViT-B \cite{ViT}, and self-distillation for DINO-base \cite{dino}). For both models, we extract patch tokens of the last block and reorder them to construct feature maps, creating $\mathbf{z} \in \mathbb{R}^{768 \times 16 \times 16}$.

\begin{table}[t]
  \centering
  \small
  \caption{\textbf{Dataset statistics} used in our experiments. ‘Res.’ represents image resolution, where $H=W$. ‘N’ and ‘A’ denote the numbers of normal and anomalous samples, respectively.}
  \label{tab:sup_dataset_stats}
  \begin{tabular}{lcccc}
    \toprule
    & \textbf{MVTecAD} 
    & \textbf{VisA} 
    & \textbf{MPDD}
    & \textbf{BMAD} \\
    \midrule
    \textbf{Classes}        & 15    & 12   & 6    & 6 \\
    \textbf{Domain}         & Industrial & Industrial & Metal &  Medical \\
    \textbf{Res.}     & 1024 & 256 & 1024 & 1024 \\
    \textbf{Train (N)}      & 2,629 & 8,659 & 888  &  52742 \\
    \textbf{Test (N)}       & 467   & 962  & 176  &  4540 \\
    \textbf{Test (A)}       & 1,258 & 1,200 & 282  &  22632 \\
    \bottomrule
  \end{tabular}
\end{table}

\begin{table}[t]
  \centering
  \small
  \caption{\textbf{Hyperparameters} of diffusion modeling.}
  \label{tab:sup_configurations}
  \begin{tabular}{@{}ll@{}}
    \toprule
    \textbf{Parameter}                       & \textbf{Value}                                \\ 
    \midrule
    \multicolumn{2}{@{}l}{\textbf{Diffusion Settings}} \\
    \midrule
    Training diffusion steps ($T$)           & 1000                                          \\
    Inference diffusion steps ($S$)          & 3                                             \\
    Noise scheduler                          & Linear, $\beta_1 = 1 \times 10^{-4}$, $\beta_T = 0.02$ \\
    Parameterization                         & $\epsilon$-prediction                         \\
    Noise scale ($\sigma$)                   & Fixed                                         \\
    \midrule
    \multicolumn{2}{@{}l}{\textbf{Training Settings}} \\
    \midrule
    Batch size                               & 8                                             \\
    Optimizer                                & AdamW                                         \\
    Gradient clipping threshold              & 1.0                                           \\
    Initial learning rate                    & $1\times10^{-6}$                              \\
    Peak learning rate                       & $5\times10^{-5}$                              \\
    Final learning rate                      & $5\times10^{-6}$                              \\
    Learning rate scheduler                  & Warmup + cosine decay                         \\
    Warmup epochs                            & 40                                            \\
    Total number of epochs                   & 300                                           \\
    Weight decay                             & 0                                             \\
    \bottomrule
  \end{tabular}
\end{table}

\subsection{Evaluation Pipeline}
We evaluate AD performance at both image- and pixel-levels using the following metrics:
  \begin{itemize}
    \item \textbf{Area Under the Receiver Operating Characteristic Curve (AU-ROC)} provides an intuition on how well the model takes a trade-off between recall and precision in a threshold-independent manner. 
    \item \textbf{Average Precision (AP)} conducts a more practical evaluation than AU-ROC, where the test set contains far fewer anomalous samples compared to normal ones.
    \item \textbf{Maximum F1-score ($\boldsymbol{\mathrm{F1}_{\max}}$)} measures the F1 score on the best-performed threshold, which reflects actual performance on the real problem. 
  \end{itemize}
To assess pixel-level anomaly localization, we employ:
  \begin{itemize}
    \item \textbf{Area Under the Per-Region-Overlap Curve (AU-PRO)} calculates the IoU between prediction and anomalous region, which is more robust to the small anomalies \cite{MVTecAD}. 
  \end{itemize}
% We utilize anomalib \cite{anomalib}, which is a well-known AD library, to calculate these metrics. 

\subsection{Experimental Environment}
Our implementations are based on Pytorch-2.7.1 \cite{torch} and Python-3.10. We show the machine specification for the training and evaluation in our experiments in \Cref{tab:sup_machines}. 

\begin{table}[t]
  \centering
  \small
  \caption{\textbf{Computing environment} in our experiments. }
  \label{tab:sup_machines}
  \begin{tabular}{lll}
    \toprule
    \textbf{Component} & \textbf{Training} & \textbf{Evaluation} \\
    \midrule
    CPU & Xeon 8468 ×2 & i9-14900KF \\
    GPU & NVIDIA H200 ×8 & RTX 4090 (24GB) \\
    RAM & 1.0 TiB ($\approx$1024GB) & 31 GB \\
    OS  & RHEL 9.4 (Plow) & Ubuntu 24.04.2 LTS \\
    \bottomrule
  \end{tabular}
\end{table}

To ensure reproducibility, we fixed random seeds for all libraries (NumPy, PyTorch, etc.) and disabled nondeterministic GPU operations if possible.

It is important to understand that we do not train any extra conditioning modules nor perform reconstruction, just train unconditional diffusion models in the standard way. This leads to significant training time reduction (e.g., for MVTecAD training time, InvAD: 2.5h on 8×H200 @28 GB/GPU—~2 × faster than DiAD \cite{DiAD}’s 5h). The use of 8× H200 is optional, \textit{not required}. Specifically, training can be performed on GPUs with more than 32 GB of memory for our largest models.

\section{Addtional ablations}
\label{sec:addtional_ablation}
\textbf{Inversion schedule.} 
To accelerate inversion computation, we uniformly set $S=3$ to skip intermediate steps within the original $T=1000$ steps, and create the subset of timesteps $\tau_3 = [333,666,999]$ (\ie, uniform scheduler). We include an ablation by varying the scheduling policy $g(u)$ for examining the impact of alternative schedules. The numerical results are shown in \Cref{tab:sup_inv_schedule_ablation}. We can see that the accuracy across all schedules is mostly consistent, with the uniform schedule performing best for $S$=3. We thus adopt this commonly used uniform schedule in our implementation.

\textbf{Feature resolutions.} 
To utilize the high-level semantic features of backbones, we follow the latent diffusion manner by extracting the feature maps $\mathbf{z} \in \mathbb{R}^{C\times h \times w}$ and operating diffusion modeling on $\mathbf{z}$. While this contributes to the improvements in both accuracy and inference speed, backbones may overly compress the anomalous information, especially for small $h\times w$. In \Cref{tab:sup_feature_res_ablation}, we conduct a design ablation on feature resolution ($h\times w$) and report MVTecAD mAD by different anomaly sizes (anomalous pixel ratio) for inspecting optimal feature resolution. The default 16$\times$16 is best overall (83.70) and on Tiny/Small (78.54/80.65), exceeding both 8$\times$8 and 24$\times$24 settings. This suggests performance is not simply limited by upsampling, and increasing resolution is not monotonically beneficial; there is an optimal resolution–efficiency balance.

\begin{table}[t]
  \centering
  \small
  \caption{\textbf{Inversion schedule ablation} under different total diffusion steps $S$ in multi-class MVTecAD with mAD.}
  \label{tab:sup_inv_schedule_ablation}
  \begin{tabular}{lccc}
    \toprule
    \textbf{$g(u),\, u = t/T$} & \textbf{$S = 3$} & \textbf{$S = 10$} & \textbf{$S = 100$} \\
    \midrule
    Uniform $g(u)=u$ & \textbf{83.35} & \textbf{82.11} & \textbf{79.35} \\
    Quad $g(u)=u^2$ & 82.37 & 80.76 & 79.14 \\
    Cube $g(u)=u^3$ & 81.68 & 80.84 & 79.18 \\
    Exp $g(u)=\frac{e^{5 u}-1}{e^5-1}$ & 81.48 & 80.85 & 79.23 \\
    \bottomrule
  \end{tabular}%
\end{table}

\begin{table}[t]
  \centering
  \small
  \caption{\textbf{Feature resolution ablation} under different anomaly sizes in multi-class MVTecAD with mAD.}
  \label{tab:sup_feature_res_ablation}
    \begin{tabular}{lccccc}
      \toprule
      \textbf{$h\times w$} & \textbf{Tiny} & \textbf{Small} & \textbf{Medium} & \textbf{Large} & \textbf{All} \\
      \midrule
      8$\times$8   & 73.38 & 76.38 & 79.17 & 83.57 & 82.05 \\
      16$\times$16 & \textbf{78.54} & \textbf{80.65} & \textbf{81.76} & \textbf{84.94} & \textbf{83.70} \\
      24$\times$24 & 76.33 & 78.76 & 80.00 & 81.97 & 81.44 \\
      \bottomrule
    \end{tabular}
\end{table}

\section{Extended Background of Diffusion Models}
\label{sec:sup_diffusion_background}

\subsection{Denoising Diffusion Probabilistic Models}
Denoising diffusion probabilistic models (DDPM) \cite{DDPM} are a class of latent generative models which involve progressively collapsed $T$ random variables $\mathbf{x}_1, \mathbf{x}_2, \ldots, \mathbf{x}_T$. Let us assume we have \textit{i.i.d.} data samples $\mathcal{X} = \{\mathbf{x}_0 \mid \mathbf{x}_0 \sim q_0(\mathbf{x})\}$ where $q_0(\mathbf{x})$ is an unknown data distribution. DDPM aims to approximate this data distribution by tracing the data collapsing process (forward process) in reverse order.
In general, the forward process $q\bigl(\mathbf{x}_{1:T}\mid \mathbf{x}_0\bigr)$ is represented as a Markov process with the Gaussian transition kernel: 
\begin{align}
q\bigl(\mathbf{x}_{1:T} \mid \mathbf{x}_0\bigr)
&:= \prod_{t=1}^T q\bigl(\mathbf{x}_t \mid \mathbf{x}_{t-1}\bigr),
\label{eq:sup_forward_process1}
\\
\text{where}\quad
q\bigl(\mathbf{x}_t \mid \mathbf{x}_{t-1}\bigr)
&:= \mathcal{N}\!\Bigl(
    \sqrt{\frac{\alpha_t}{\alpha_{t-1}}}\,\mathbf{x}_{t-1},
    \bigl(1 - \frac{\alpha_t}{\alpha_{t-1}}\bigr)\mathbf{I}
\Bigr).
\label{eq:sup_forward_process2}
\end{align}
The noise schedule is defined by the monotonically decreasing sequences $\{\alpha_t\}_{t=1}^T$ in the above equation. Notably, from the reproducibility of the Gaussian distribution, we have: 
\begin{align}
q\left(\mathbf{x}_t \mid \mathbf{x}_0\right) = \mathcal{N}\left(\mathbf{x}_t ; \sqrt{\alpha_t} \mathbf{x}_0, 
\left(1-\alpha_t\right) \mathbf{I}\right).
\label{eq:sup_q_prop}
\end{align}
Since we can efficiently compute latent variables $\mathbf{x}_t$ given a clean sample $\mathbf{x}_0$, it provides a GPU-friendly diffusion model training. 
Also, we define the generative process $p_\theta\left(\mathbf{x}_{0: T}\right)$ as the Markov process with a learnable Gaussian transition kernel: 
\begin{align}
p_\theta\left(\mathbf{x}_{0: T}\right) := p_\theta\left(\mathbf{x}_T\right) \prod_{t=1}^T p_\theta^{(t)}\left(\mathbf{x}_{t-1} \mid \mathbf{x}_t\right), 
\label{eq:sup_gen_process2}
\\
\text{where} \quad
p_\theta^{(t)}(\mathbf{x}_{t-1} \mid \mathbf{x}_t) := \mathcal{N}\bigl(\mathbf{x}_{t-1}; \boldsymbol{\mu}_\theta^{(t)}(\mathbf{x}_t), \sigma_t^2 \mathbf{I}\bigr).
\label{eq:sup_gen_process2}
\end{align}
In DDPM, the posterior mean is modeled using a noise prediction neural network $\epsilon_\theta^{(t)}(\mathbf{x}_t)$: 
\begin{align}
    \boldsymbol{\mu}_\theta^{(t)}(\mathbf{x}_t) = \frac{\sqrt{\alpha_{t-1}}}{\sqrt{\alpha_{t}}}\bigl(\mathbf{x}_t - \frac{1 - (\alpha_t/\alpha_{t-1})}{\sqrt{1 - \alpha_t}}\epsilon_\theta^{(t)}(\mathbf{x}_t)
    \bigr).
\end{align}
The diffusion objective is expressed as: 
\begin{equation}
    \mathcal{L}_\gamma\left(\epsilon_\theta\right) \coloneqq \sum_{t=1}^T \gamma_t \mathbb{E}_{\mathbf{x}_0 \sim q\left(\mathbf{x}_0\right), \epsilon_t \sim \mathcal{N}(\mathbf{0}, \mathbf{I})}\left[\left\|\epsilon_\theta^{(t)}(\mathbf{x}_t)-\epsilon_t\right\|_2^2\right],
    \label{eq:sup_diff_objective}
\end{equation}
where $\mathbf{x}_t$ is efficiently sampled by \Cref{eq:sup_q_prop}, and $\mathbf{\gamma} \coloneqq [\gamma_1, \ldots, \gamma_T]^T$ denotes the weighting vector.

\begin{figure*}[htbp] % t は top (上部配置)
    \centering
    \begin{subfigure}{0.49\textwidth}
        \centering
        \includegraphics[width=\linewidth]{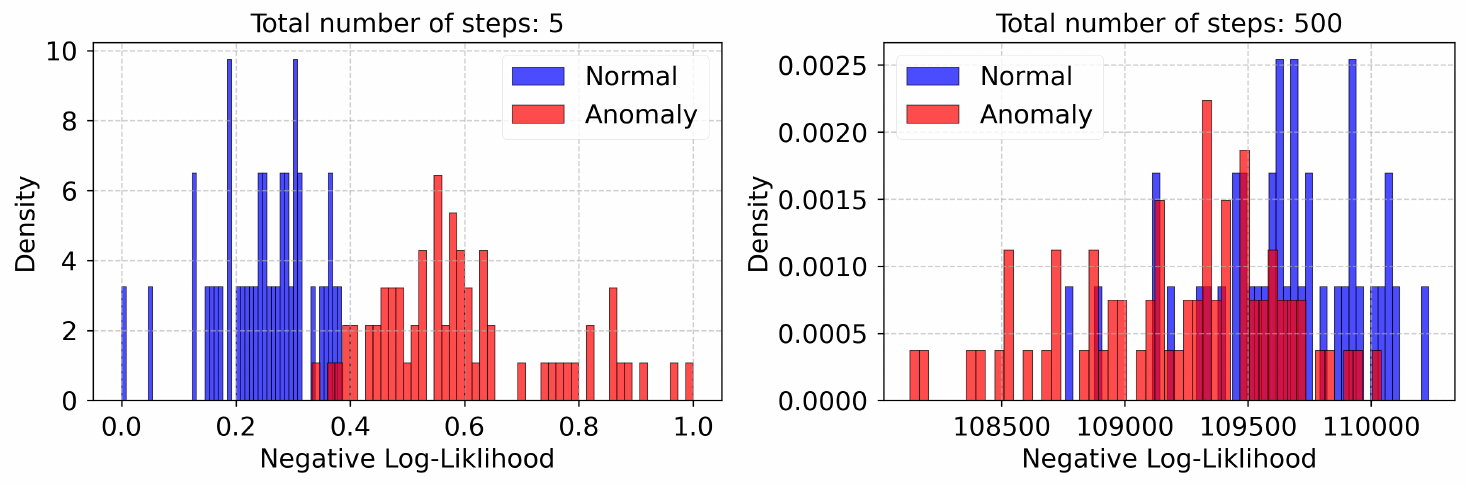}
        \caption{Histogram of normal and anomalous samples with NLL scoring.}
        \label{fig:method_recon}
    \end{subfigure}
    \hfill
    \begin{subfigure}{0.49\textwidth}
        \centering
        \includegraphics[width=\linewidth]{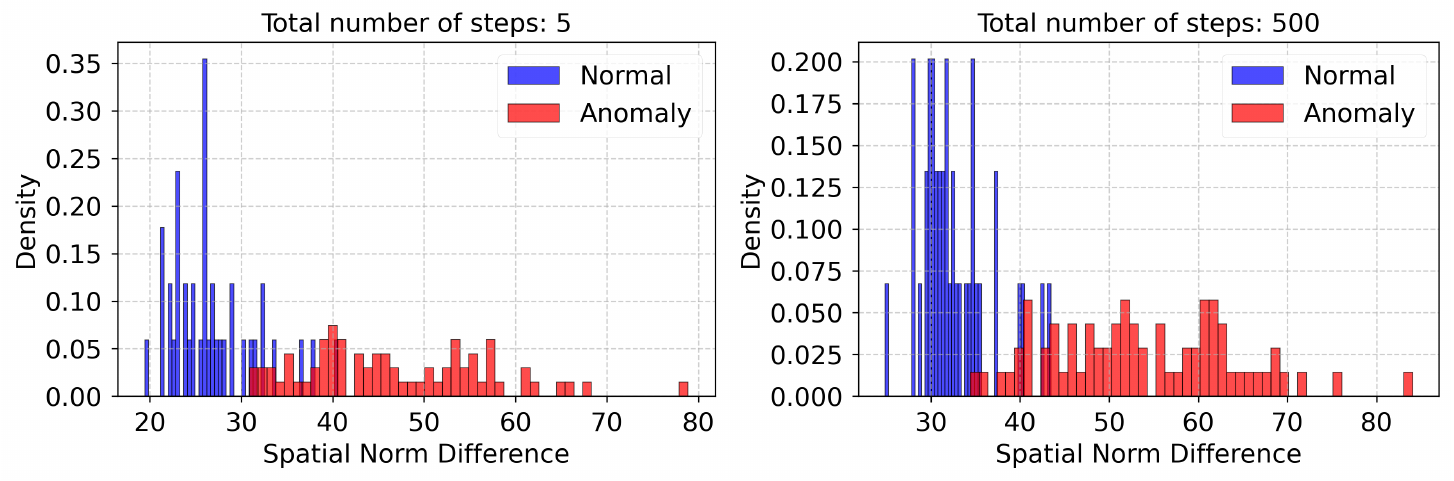}
        \caption{Histogram of normal and anomalous samples with \textit{our NLL+Diff} scoring.}
        \label{fig:method_ours}
    \end{subfigure}
    \caption{\textbf{Comparison of the histogram} of normal and anomalous samples, with the conventional NLL scoring (a) and our proposed \textit{NLL+Diff} scoring (b), on the test set of \textit{hazelnut} in MVTecAD.}
    \label{fig:sup_histo}
\end{figure*}

\subsection{Denoising Diffusion Implicit Models}
The forward process in \Cref{eq:sup_forward_process1} is a Markov process where the current state $\mathbf{x}_t$ solely depends on the previous state $\mathbf{x}_{t-1}$. 
% sampling efficiency
Denoising diffusion implicit models (DDIM) \cite{DDIM} is a non-Markov diffusion model where each forward process is conditioned on both $\mathbf{x}_0$ and $\mathbf{x}_{t-1}$. Specifically, the forward posterior is defined as: 
\begin{multline}
    q_\sigma\left(\mathbf{x}_{t-1} \mid \mathbf{x}_t, \mathbf{x}_0\right) \\= 
    \mathcal{N}\Biggl(
        \sqrt{\alpha_{t-1}} \mathbf{x}_0 +
        \sqrt{1 - \alpha_{t-1} - \sigma_t^2} \cdot 
        \frac{\mathbf{x}_t - \sqrt{\alpha_t} \mathbf{x}_0}{\sqrt{1 - \alpha_t}}, 
        \sigma_t^2 \boldsymbol{I}
    \Biggr),
    \label{eq:sup_ddim_posterior}
\end{multline}
where $\{\sigma_t\}_{t=1}^T$ controls the uncertainly of the diffusion trajectories. Since this posterior is 
designed to match $q(\mathbf{x}_t \mid \mathbf{x}_0)$ as DDPM, the generative process can be trained with the same objective described in \Cref{eq:sup_diff_objective}, for any $\{\sigma_t\}_{t=1}^T$. When $\sigma_t \rightarrow 0$, the generative process becomes deterministic and can be 
represented as: 
\begin{equation}
\mathbf{x}_{t-1}=\sqrt{\alpha_{t-1}}\boldsymbol{f}_\theta\left(\mathbf{x}_t, t\right)+\sqrt{1-\alpha_{t-1}} \epsilon_\theta^{(t)}(\mathbf{x}_t),
\label{eq:sup_ddim_gen}
\end{equation}
where $\boldsymbol{f}_\theta(\mathbf{x}_t, t) = (\mathbf{x}_t - \sqrt{1 - \alpha_t} \epsilon_\theta^{(t)}(\mathbf{x}_t))/\sqrt{\alpha_t}$. For more detailed proof and derivation of DDIM, see \cite{DDIM}. 

\subsection{Derivation of DDIM Inversion}
From \Cref{eq:sup_ddim_gen}, we have: 
\begin{equation}
    \mathbf{y}_{t-1} - \mathbf{y}_t = (p_{t-1} - p_t)\epsilon_\theta^{(t)}(\mathbf{x}_t),
    \label{eq:sup_ddim_de}
\end{equation}
where \(\mathbf{y}_t \coloneqq \mathbf{x}_t/\sqrt{\alpha_t}\) and \(p_t \coloneqq \sqrt{1/{\alpha_t} - 1}\). In the limit of $T \rightarrow \infty$, the above differential equation converges to a continuous-time ODE: 
\begin{equation}
    \mathrm{d}\mathbf{y}_t = \epsilon_\theta^{(t)} \,\mathrm{d}p_t.
    \label{eq:sup_ddim_ode}
\end{equation}
Based on the ODE formulation of DDIM, we can derive \Cref{eq:sup_ddim_gen} by applying Euler integration to \Cref{eq:sup_ddim_ode} with the step size $\Delta p_t^+ := p_{t-1} - p_t (>0)$. Similarly, by applying the forward Euler method to the \Cref{eq:sup_ddim_ode} within the range $[t-1, t]$, we have: 
\begin{equation}
    \mathbf{y}_{t} = \mathbf{y}_{t-1} + \Delta p_t^- \epsilon_\theta^{(t-1)}(\mathbf{x}_{t-1}), 
    \label{eq:sup_ddim_inversion1}
\end{equation}
where $\Delta p_t^- := p_t - p_{t-1}$. Next, replacing $\mathbf{y}_t$ with $\mathbf{x}_t/\sqrt{\alpha}_t$ in \Cref{eq:sup_ddim_inversion1} yields: 
\begin{equation}
    \mathbf{x}_{t} = \sqrt{\alpha_{t}}\bigl( \frac{\mathbf{x}_{t-1}}{\sqrt{\alpha_{t-1}}} + \Delta p_t^- \epsilon_\theta^{(t-1)}(\mathbf{x}_{t-1})\bigr).
    \label{eq:sup_ddim_inversion2}
\end{equation}

It can be easily noticed that Equation \Cref{eq:sup_ddim_inversion2} is essentially equivalent to Equation (6). A similar discussion applies to the subset of timesteps $\bm{\tau}_S = [\tau_1, \tau_2, \ldots, \tau_S=T] \subset \{1, 2, \ldots, T\}$ by considering arbitray time interval $[\tau_s, \tau_{s+1}]$. 

\section{Reverse Scoring Problem}
\label{sec:reverse_scoring}

As shown in \Cref{tab:ablation_scoring,tab:ablation_step} (in the main paper), we observe a performance drop when the number of diffusion steps \( S \) increases in conjunction with NLL-based anomaly scoring. To investigate the underlying cause of this phenomenon, we visualize the NLL distribution across different values of \( S \) in \Cref{fig:sup_histo}. Interestingly, the histogram of NLL exhibits counterintuitive results: as \( S \) increases by 500, normal images are assigned lower likelihoods, suggesting that learned normal samples occupy lower-density regions than anomalous ones. This observation aligns with findings in flow-based out-of-distribution (OOD) detection literature \cite{DoGMKnow,OODTP}, where a similar tendency arises due to the nature of high-dimensional data distributions. This is known as the \textit{reverse-scoring} issue, which is widely acknowledged in dealing with high-dimensional space.

\end{document}